\documentclass[11pt]{article}

\usepackage[final]{acl}

\usepackage{times}
\usepackage{latexsym}

\usepackage[T1]{fontenc}

\usepackage[utf8]{inputenc}

\usepackage{microtype}

\usepackage{inconsolata}

\usepackage{graphicx}

\usepackage{subcaption}
\usepackage{enumitem}
\usepackage{graphicx}
\usepackage[table]{xcolor}
\definecolor{psblue}{RGB}{220,235,255}
\definecolor{rdorange}{RGB}{255,235,210}
\usepackage{amsmath}
\usepackage{caption}
\usepackage{multirow}
\usepackage{booktabs}
\usepackage{tabularx}
\usepackage{array}
\usepackage{ragged2e}
\usepackage{placeins}
\newcolumntype{Y}{>{\RaggedRight\arraybackslash}X}

\title{PSI-Bench: Interpretable and Clinically Meaningful Evaluation of Depression Patient Simulators}

\author{
  \textbf{Nguyen Khoi Hoang}$^{1,2}$ \quad
  \textbf{Shuhaib Mehri}$^1$ \quad
  \textbf{Tse-An Hsu}$^1$ \quad
  \textbf{Yi-Jyun Sun}$^1$ \quad \\
  \textbf{Quynh Xuan Nguyen Truong}$^3$ \quad 
  \textbf{Khoa D Doan}$^2$ \quad 
  \textbf{Dilek Hakkani-Tür}$^1$ \\[10pt]
  $^1$University of Illinois Urbana-Champaign, \\
  $^2$VinUni-Illinois Smart Health Center,\\ $^3$College of Health Sciences, VinUniversity \\[5pt]
  \texttt{\{nkhoang2, dilek\}@illinois.edu}
}

\begin{document}
\maketitle

\begin{abstract}
Patient simulators are gaining traction in mental health training by providing scalable exposure to complex and sensitive patient interactions. Simulating depressed patients is challenging, as safety constraints and high patient variability complicate simulations and underscore the need for simulators that capture diverse and realistic patient behaviors. However, existing evaluations heavily rely on LLM-judges with poorly specified prompts and do not assess behavioral diversity. We introduce \textbf{PSI-Bench}\footnote{We release the code and simulated conversation datasets at: \url{https://github.com/nguyenkhoihoang/PSI-Bench}}, an automatic evaluation framework that provides interpretable, clinically meaningful diagnostics of depression patient simulator behavior across turn-, dialogue-, and population-level dimensions. Using PSI-Bench, we benchmark seven LLMs across two simulator frameworks and find that simulators produce overly long, lexically diverse responses, show reduced variability, and move through therapeutic
stages and toward positive valence too quickly. We also show that the simulation framework has a larger impact on fidelity than the model scale. Results from a human study demonstrate that our benchmark is strongly aligned with judgments of mental health professionals. Our work reveals key limitations of current depression patient simulators and provides an interpretable, extensible benchmark to guide future simulator design and evaluation.
\end{abstract}

\section{Introduction}

Large Language Models (LLMs) are increasingly used to simulate human behavior in interactive settings where real-world practice is costly, sensitive, or difficult to scale \citep{park2023generative, yao2024tau, zhou2024sotopia}. In mental health education, LLM-based patient simulators can generate clinically plausible responses and act as stand-ins for real patients, offering a practical training environment for novice clinicians \citep{liu-etal-2025-eeyore, wang-etal-2024-patient, louie-etal-2024-roleplay}. The fidelity of these simulators is critical, as divergence from human patient communication, such as in language use or conversational progression, can risk reinforcing misconceptions and undermining training effectiveness \citep{mehri2025goal, abdulhai2025consistently, naous2026flipping}. Therefore, it is essential to evaluate simulator fidelity with human patient behavior. This work focuses on a particularly challenging domain of mental health: depression. Simulating depression is difficult as LLMs' safety training limits their ability to mimic depressive speech patterns \citep{liu-etal-2025-eeyore} and as depression varies widely between individuals \citep{fried2015depression}.

\begin{figure*}[t]
    \centering
    \includegraphics[width=\linewidth]{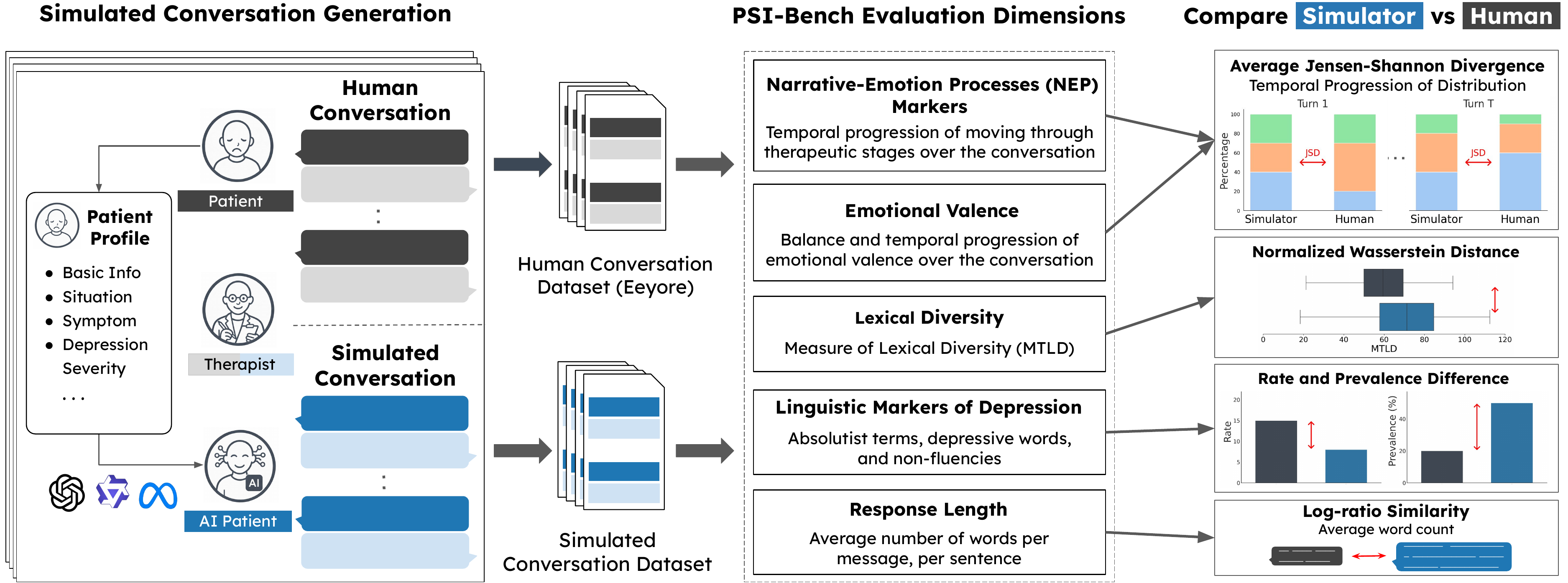}
    \caption{Overview of PSI-Bench. From the Eeyore dataset \cite{liu-etal-2025-eeyore} of depression-related human patient conversations, we condition the patient simulator on the patient profiles derived from these conversations to generate a simulated conversation dataset. Both the human and simulated datasets are then evaluated independently using the PSI-Bench dimensions, and the resulting statistics are compared.}
    \label{fig:overview}
    \vspace{-10pt}
\end{figure*}

Current automatic evaluations of simulator fidelity rely on LLM-judges to assess how well generated responses align with the target patient profiles \citep{liu-etal-2025-eeyore, wang-etal-2024-patient}. They score responses on a 5-point Likert scale \citep{likert1932technique} using generic prompts, entrusting the LLM-judge to infer nuanced clinical distinctions without explicit grounding in psychological theory or human patient data. This setting may not be reliable, as \citet{wang-etal-2024-patient} reports opposite trends between LLM-judge fidelity scores and expert ratings. Moreover, it lacks the interpretability needed to reveal \textit{where} a simulator diverges from human patient behavior, offering little guidance for improvement. Additionally, existing evaluation approaches \citep{sabour2026patienthub, wang-etal-2024-patient, liu-etal-2025-eeyore} report an average score without examining whether the simulators reflect the distributional diversity of patient populations. Patients can exhibit varying linguistic and interactional patterns \cite{ruiz2025role}, especially among depressed patients \cite{fried2015depression}. Simulators that fail to reflect such diversity may introduce bias in clinician training \cite{schwieters2023default}.

Recognizing these limitations, we propose \underline{P}atient-\underline{SI}mulator-Bench (\textsc{PSI-Bench}), a framework for interpretable, clinically meaningful evaluation of depression simulators. PSI-Bench comprises five dimensions supported by established psychological and psycholinguistic research \citep{rude2004language, trifu2024linguistic,10.1371/journal.pone.0251787,tolboll2019linguistic, smirnova2018language,beck1967depression}, ensuring that the evaluation meaningfully captures the characteristics of depression behavior. Given a set of reference human conversations and patient profiles derived from them, we evaluate generated and human responses along these dimensions to quantify simulator alignment with human patient behavior. Due to privacy and ethical constraints on accessing real clinical data, we apply PSI-Bench to Eeyore \cite{liu-etal-2025-eeyore}, an expert-curated English corpus of crowd-sourced and acted depression-related conversations paired with extracted patient profiles. Such proxy interactions are well-established in mental health education \citep{dawood2024evaluation}, and the framework can be applied to authentic conversation datasets as they become available. To assess the reliability and diagnostic value of PSI-Bench, we use PSI-Bench to investigate three research questions:

\begin{itemize}[leftmargin=10pt,topsep=0pt, itemsep=0pt]
\item \textbf{RQ1:} How do responses generated by current depression patient simulators differ from human reference across turn-level, dialogue-level, and population-level?
\item \textbf{RQ2:} How do different LLMs and simulation strategies affect simulator fidelity?
\item \textbf{RQ3:} Do the findings from answering RQ1 and RQ2 using PSI-Bench align with judgments of mental health professionals?
\end{itemize}

We evaluate seven LLMs across two simulation frameworks \citep{wang-etal-2024-patient, louie-etal-2024-roleplay}, and reveal key limitations not captured by previous automatic evaluation approaches. At the population level, simulators show less variability and diverge from human reference in dialogue progression: they move through therapeutic stages more rapidly and follow a uniform negative-to-positive valence trajectory. They also produce overly long, lexically diverse responses at the message level, and exhibit a more spread out presence of linguistic markers of depression at the dialogue level.

We aggregate scores across all dimensions into a unified alignment score to rank the simulators (Table~\ref{tab:overall_benchmark}), while retaining interpretable diagnostics of each evaluation dimension. We find that the simulation framework can have a larger impact  than LLM capability, and smaller models can be closer to human patient communication than larger ones. Finally, we validate our benchmark via a human study with 20 mental health professionals. The results show near-perfect agreement between their pairwise preferences and PSI-Bench's ordering ($\kappa = 0.82$), supporting PSI-Bench as a reliable evaluation framework. Our main contributions are:

\begin{itemize}[leftmargin=12pt,topsep=0pt, itemsep=0pt]
    \item We introduce PSI-Bench, an interpretable, clinically meaningful benchmark for evaluating patient simulators by alignment with human reference, showing strong agreement with judgments of mental health professionals.
    \item Using PSI-Bench, we highlight previously underexplored simulator divergences from human patient behavior, and quantify the impact of simulation framework, scale, and model family on simulator fidelity.
\end{itemize}


\section{Related Work}
\label{sec:related_work}
\subsection{LLM-based Simulators}

\noindent \textbf{User Simulators for Dialogue Systems} User simulators have been studied since the mid-1990s to address the cost and difficulty of collecting real conversational data, particularly in specialized domains \citep{Pietquin_Hastie_2013}. While traditional rule-based, statistical, and early neural approaches struggled to capture natural human behavior, LLMs offer powerful language capabilities to overcome these limitations \citep{naveed2025comprehensive}. Existing work mainly explores two applications of LLM-based simulators. First, they are used to \textit{train and evaluate AI systems} by generating dialogue data \citep{finch2024diverse, bao2023synthetic} or as interactive environments for online reinforcement learning \citep{carta2023grounding}. Second, they are used as \textit{training tools for human learners} in domains where practicing with real individuals is costly or sensitive, such as nurse communication training \citep{lee2025adaptive} and counseling \citep{na2025survey}.

\vspace{0.5em}

\noindent \textbf{LLM-based Patient Simulators in Mental Health} Within the counseling domain, several works \citep{chen-etal-2023-soulchat, lee-etal-2024-cactus} focus on the first application -- generating counseling datasets to improve counselor LLMs, where a patient simulator's design and evaluation are less prioritized. In contrast, our evaluation framework targets the second application -- simulating patients for training novice counselors, where the central emphasis is on designing simulated patients with realistic behaviors \citep{liu-etal-2025-eeyore,wang-etal-2024-patient,louie-etal-2024-roleplay}. 

\subsection{Evaluation of Patient Simulators}
\noindent \textbf{Evaluating User Simulators} Before the advent of LLMs, \citet{Pietquin_Hastie_2013} grouped the evaluation of user simulators into two groups: \textit{turn-level} metrics that evaluate individual responses (e.g., precision and recall) and \textit{dialogue-level} metrics that evaluate general interaction (e.g., task completion). Though fast and reproducible, these traditional metrics rely on surface lexical overlap and miss deeper nuances \citep{gu2025surveyllmasajudge}. Along with the rise of LLM-based simulators, LLM-based evaluation methods emerged with their strong reasoning and language capabilities, enabling more context-aware assessment but also introducing new systematic bias \citep{gu2025surveyllmasajudge, gallegos-etal-2024-bias}. Evaluation can also be framed as \textit{intrinsic} versus \textit{extrinsic}: intrinsic methods assess simulator outputs directly (e.g., realism and diversity), while extrinsic methods measure downstream task impact. \citet{naous2026flipping} show that extrinsic evaluation alone can be misleading as over-cooperative LLM users may inflate system performance, motivating greater emphasis on intrinsic methods.
\vspace{0.5em}

\noindent \textbf{Evaluating LLM-based Patient Simulators in Counseling} Simulation systems aimed at training or benchmarking counselor LLMs \citep{lee-etal-2024-cactus, wang2025care} are often automatically evaluated using extrinsic metrics of counseling effectiveness. However, consistent with the findings of \citet{naous2026flipping}, such simulated patients can be implicitly cooperative and demonstrate rapid improvement \citep{abdulhai2025consistently}, thus increasing extrinsic performance scores.

Patient simulators for training novice counselors are typically evaluated for alignment, realism, and training utility \citep{liu-etal-2025-eeyore, wang-etal-2024-patient, louie-etal-2024-roleplay, sabour2026patienthub}. Expert human evaluation assess profile alignment, behavioral fidelity, and pedagogical effectiveness \citep{wang-etal-2024-patient}. While it is often impractical to use expert evaluation alone as it is costly and not scalable, current automatic evaluation pipelines which rely on LLM-Judge Likert ratings are reported to be unreliable or even contradict expert judgments \citep{wang-etal-2024-patient}. Additionally, they typically report an average score without examining broader behavioral distributions reflecting real-world clinical variability. Our evaluation framework addresses these limitations.

\section{PSI-Bench} \label{sec:evaluation_dimensions}
Figure \ref{tab:overall_benchmark} provides an overview of PSI-Bench. We use the reference conversation dataset’s pre-extracted patient profiles by default; if a simulator provides its own profile-extraction method, we instead derive the patient representations from the reference dataset using that method. We generate simulated conversations by conditioning patient simulators on the patient profiles and pairing them with an LLM therapist. Both human reference and synthetic datasets are then evaluated independently using PSI-Bench dimensions, and the resulting statistics are compared. 

Our reference dataset combines three dyadic conversation datasets from Eeyore \cite{liu-etal-2025-eeyore}, curated from public depression-related dialogue sources by extracting depressive traits to guide data filtering and construct structured psychological profiles paired with each conversation (see Appendix \ref{sec:real_dataset}). While the conversations contain crowd-sourced and acted interactions, such proxy interactions are commonly used in mental health education via standardized patients due to privacy constraints \cite{dawood2024evaluation}, and instruction-tuning on these conversations improved simulator realism \cite{liu-etal-2025-eeyore}. PSI-Bench can be readily extended to datasets containing authentic clinical conversations as such resources become available.

We ground our assessment in behaviors observed in depression-related communication and analyze simulator behavior across five interpretable dimensions. \textbf{Section \ref{sec:dimensions}} introduces and independently applies these dimensions to reference and synthetic conversations. \textbf{Section \ref{sec:method_benchmark}} describes how simulators' scores are compared relative to human patient scores at multiple levels: \textit{turn-level} (e.g., response length), \textit{conversation-level} (e.g., progression and valence change), and \textit{population-level} (e.g., diversity across profiles), capturing realistic patterns and variation. Each dimension is aggregated into normalized scalar scores for comparison, ranking, and correlation with human evaluation.

\subsection{Evaluation Dimensions}\label{sec:dimensions}

\noindent \textbf{Narrative-Emotion Processes (NEP) Markers} are standardized categories rated by observers that capture how patients manifest narrative content and emotional experience during psychotherapy \citep{Angus04052017}. Because the evolving pattern of these markers over a session reflects therapeutic progress, they provide a basis for assessing whether simulated patients replicate the change dynamics observed in real clinical conversations. The markers are grouped into three categories:
\textit{Problem}: repetitive, “stuck” storylines with limited reflection; \textit{Transition}: more coherent and reflective storytelling as clients begin to examine their thoughts and emotions;  \textit{Change}: clear shifts in emotional response and thought patterns, where clients may show a new understanding of key events in their life.  To track these patterns, we use \texttt{gpt-oss-120b} \citep{openai2025gptoss120bgptoss20bmodel} to classify each patient turn in conversations as Problem, Transition, Change, or Filler (neutral procedural replies without meaningful therapeutic content), using explicit definitions and decision criteria (Appendix \ref{app:ptc_classification_prompt}). 

\vspace{0.5em}

\noindent \textbf{Emotional Valence} \citet{tolboll2019linguistic, rude2004language} show that depression correlates with increased negative and decreased positive emotion words. We therefore investigate whether simulated and human patients differ in how their emotional valence manifests and evolves over the course of the conversation. We use \texttt{gpt-oss-120b} to classify each patient message as Positive, Negative, or Neutral (Appendix \ref{app:valence_classification_prompt}). We validate \texttt{gpt-oss-120b}'s NEP and valence classifications against annotations from 20 mental health professionals and find high agreement that supports its classification reliability (Section \ref{sec:human_study}). 
\vspace{0.5em}

\noindent \textbf{Lexical Diversity}
\citet{smirnova2018language} showed that patients with depression often exhibit repetition in words or ideas. We evaluate whether simulators can mimic this behavior despite LLMs’ tendency to generate highly diverse text \citep{verbose_emails_2025}. We also investigate how lexical diversity varies across human patients and whether the same LLM conditioned on different profiles can mirror such variation. We compute conversation-level lexical diversity by concatenating all patient messages in a conversation and applying the Measure of Textual Lexical Diversity (MTLD)~\citep{mccarthy2010mtld}. MTLD is defined as the mean length of word sequences required to reach a fixed ratio ($0.72$) of unique words to total words (Appendix \ref{sec:app_mtld_verbosity}). Additionally, we concatenate all patient messages across profiles into a single corpus and compute its MTLD. A lower corpus-level MTLD than the average per-conversation MTLD would suggest some degree of shared vocabulary across patients. 

\vspace{0.5em}
\noindent \textbf{Response Length}
Patients with depression tend to produce shorter utterances and more limited content in conversation \citep{trifu2024linguistic, smirnova2018language}, while LLMs tend to be more verbose \citep{verbose_emails_2025}. We assess response length by computing average number of words per message and average number of words per sentence.

\vspace{0.5em}
\noindent \textbf{Linguistic Markers of Depression}
Psychological research identifies consistent linguistic patterns in depressed patients that correlate with increased use of: (1) \textit{absolutist terms}: words that may reflect absolutist thinking underlying many cognitive distortions (e.g. "always", "never") \citep{doi:10.1177/2167702617747074}, (2) \textit{depressive words} associated with low mood and hopelessness \citep{info:doi/10.2196/17650}, and (3) \textit{non-fluencies}: hesitations and fillers that may signal cognitive or emotional strain \citep{10.1371/journal.pone.0251787}. Table~\ref{tab:depression_marker_lexicon} details the complete lexicon used. These markers enable a structured comparison of depressive linguistic tendencies between humans and simulators. We analyze the proportion of messages containing at least one marker (marker prevalence) and the rate of marker occurrences per 1{,}000 tokens (marker rate).


\begin{figure}[t]
    \centering
    \begin{subfigure}{\linewidth}
        \centering
        \includegraphics[width=\linewidth]{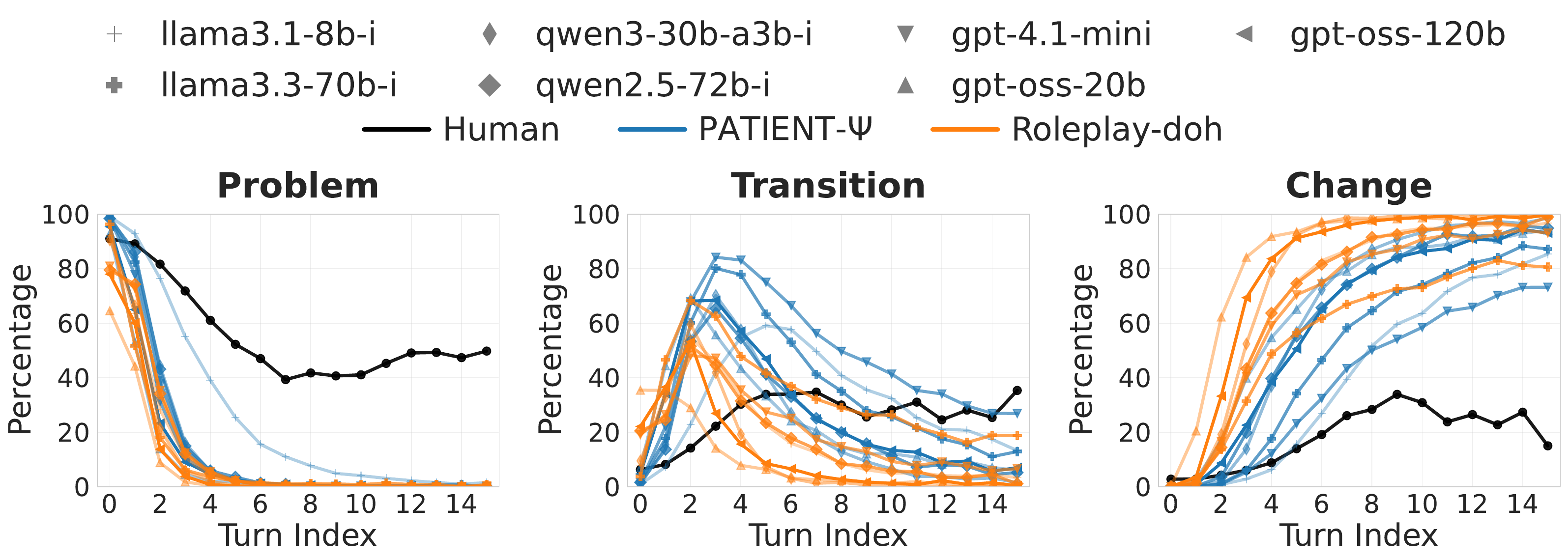}
        \vspace{-10pt}
        \caption{Population-level distribution of narrative-emotion markers. Simulators show a more rapid progress than human reference.}
        \label{fig:ptc_progression}
    \end{subfigure}
    
    \vspace{0.5em}
    
    \begin{subfigure}{\linewidth}
        \centering
        \includegraphics[width=\linewidth]{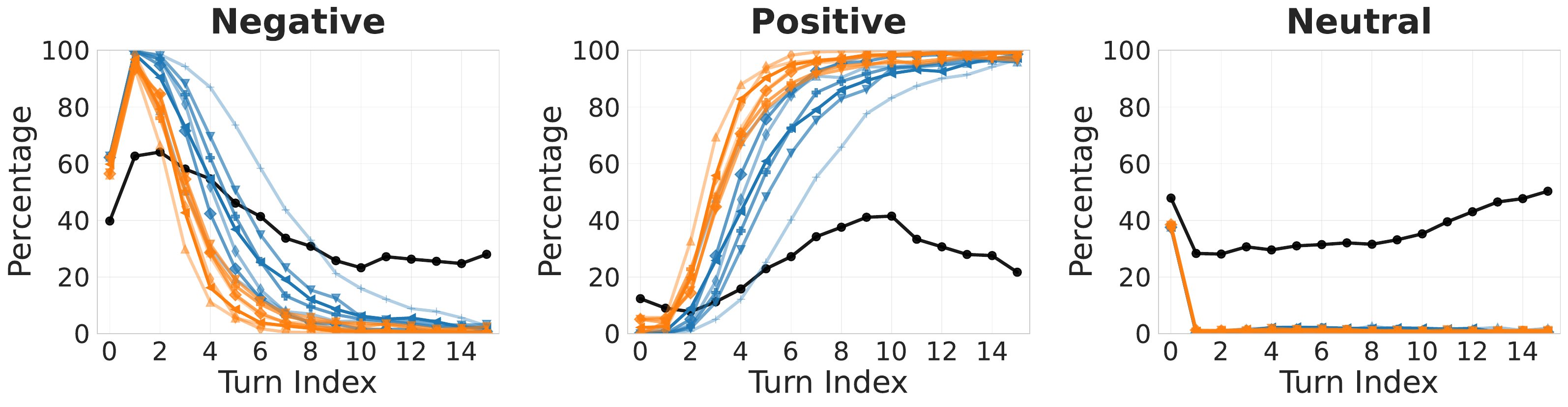}
        \vspace{-10pt}
        \caption{Population-level distribution of emotional valence. Human reference shows more neutral and varied valence, while simulators show a consistent negative-to-positive progression.}
        \label{fig:emotions_progression}
    \end{subfigure}
    \vspace{-15pt}
    \caption{Population-level progression of patient responses across turns.  Each point represents the percentage of messages assigned to a category at a given turn. }
    \label{fig:ptc_emotion_progression}
    \vspace{-10pt}
\end{figure}
\subsection{Comparative Evaluation}\label{sec:method_benchmark}
We compare simulator scores with human reference statistics to quantify alignment with human patient behavior. All metrics are converted to $0$--$100$ similarity scale, where higher values indicate higher alignment. We convert distance or divergence measures $d$ to similarity scores: $100 - d$. We denote simulator statistics by $\psi$ and reference by $r$.

\vspace{0.5em}

\noindent \textbf{Divergence of progression dynamics} For NEP markers and valence, we compare the temporal progression of label distributions. To ensure comparability, we restrict the analysis to the first $T=16$ turns for both human and synthetic conversations, yielding aligned turn-level distributions. For each turn $i$, we compute a population-level distribution on labels by aggregating all patient messages at that turn across profiles.
This produces a sequence of $T$ turn-level distributions $(\psi_i)_{i=1}^T$ for a simulator and $(r_i)_{i=1}^T$ for the reference patient population, where each $\psi_i$ and $r_i$ is a distribution on labels at turn $i$. This formulation enables us to track how the proportions of the labels evolve over the conversation. To quantify the divergence in progression dynamics between a simulator and the reference human population, 
we compute the symmetric Jensen-Shannon divergence (JSD) \cite{menendez1997jensen} at each turn and average across all turns:
\begin{equation}
\mathrm{d}
= \frac{1}{T} \sum_{i=1}^{T}
\mathrm{JSD}(\psi_i \,\|\, r_i)
\end{equation}
\noindent \textbf{Normalized Wasserstein distance} For lexical diversity, we compare the distribution of per-conversation MTLD scores between simulator and reference using Wasserstein distance \cite{villani2009wasserstein}, which captures difference in both distribution shape and spread. We convert this distance into a similarity measure and apply min–max normalization.

\vspace{0.5em}
\noindent \textbf{Log-ratio similarity score} As simulated patients are more verbose than human patients by large factors, we use log scale to account for this disparity. We measure response length alignment using a log-ratio similarity over average words per message and per sentence. For each average word count $c$, we compute:
\begin{equation}
s(c_{\psi}) = 100 \cdot \exp\left(- \left| \log \frac{c_{\psi}}{c_{\text{r}}} \right| \right).
\end{equation}
The final response length similarity score is the average of the two similarity scores.

\vspace{0.5em}
\noindent \textbf{Difference from human statistics} For linguistic markers of depression, we measure the deviation from human patient statistics using the relative (percentage) difference for the marker rate and the absolute difference for the marker prevalence, as it is already a bounded percentage. The final distance $d$ is the average of the two.

\begin{figure}[t]
    \centering
    \captionsetup{font=small}
     \includegraphics[width=\linewidth]{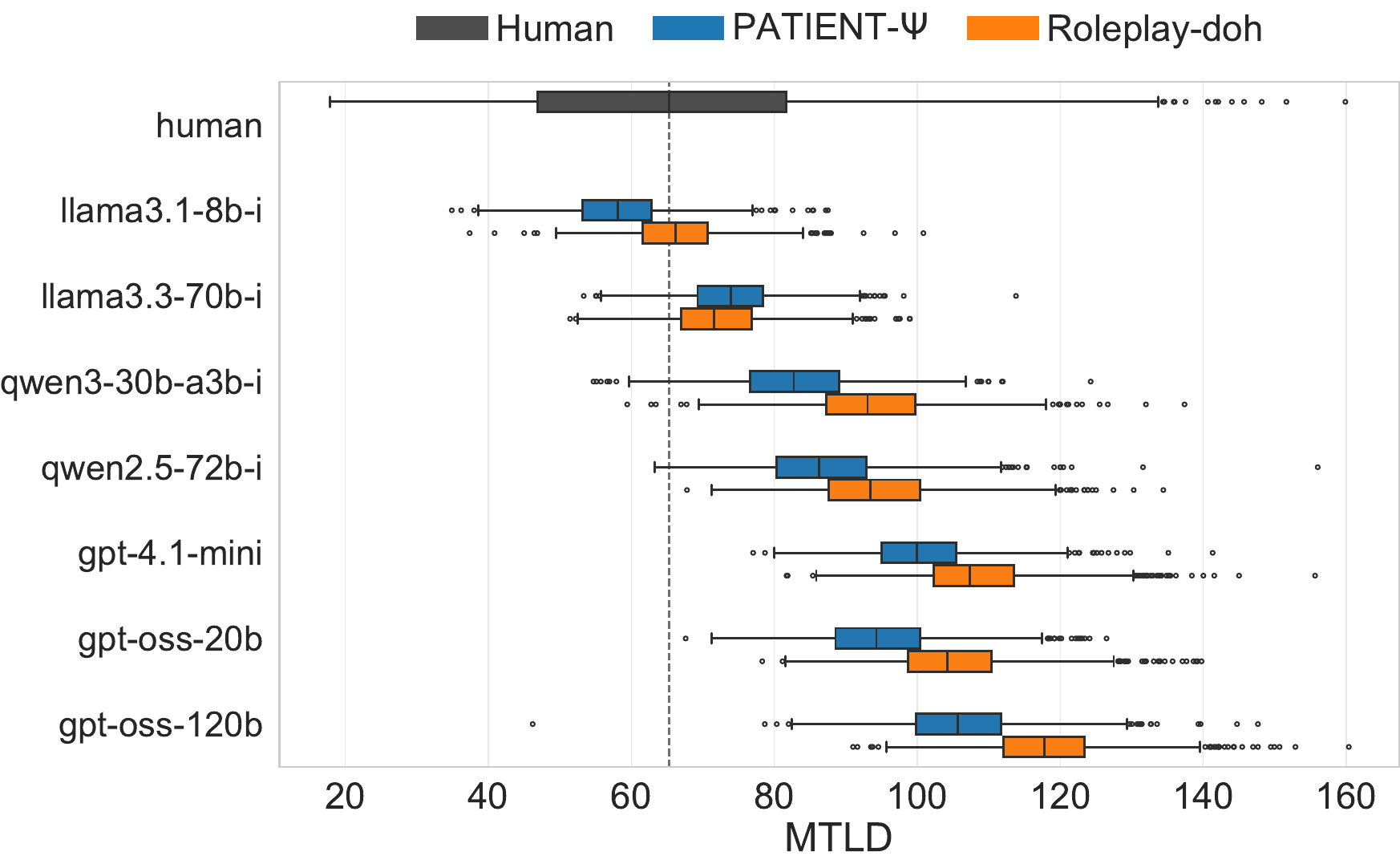}
     \vspace{-5pt}
    \caption{Boxplots of lexical diversity of human and simulated patient conversations. Simulators consistently show higher and more uniform lexical diversity than human patients.}
    \label{fig:mtld_main}
    \vspace{-10pt}
\end{figure}
\section{Experimental Setup}

We evaluate two main counsellor-training patient simulation frameworks, instantiated with one proprietary and six open-source LLMs spanning three model families (Table \ref{tab:model_alias}), for a total of 14 framework-LLM pairings. We use the same counsellor LLM \texttt{gpt-oss-20b} \citep{openai2025gptoss120bgptoss20bmodel} across all settings. Generation is capped at 16 turns or the number of patient turns in the corresponding reference conversation, whichever is smaller.  We repeat the benchmark with a second full generation run under identical settings, and another with \texttt{PsyCoPref-Llama3-8B} \citep{zhang2026preference} as the counsellor LLM. The \textbf{findings in Section \ref{sec:results} are consistent across all three runs}.

\vspace{0.5em}
\noindent \textbf{PATIENT-$\Psi$}~\citep{wang-etal-2024-patient} is a cognitive behavioral therapy--based patient simulation framework that constructs diverse patient \textit{cognitive models} -- a structured and interconnected representation of thoughts, beliefs, emotions, and behaviors. 

\vspace{0.5em}
\noindent \textbf{Roleplay-doh}~\citep{louie-etal-2024-roleplay} 
is a human-LLM collaborative framework that enables experts to customize patient simulators by converting qualitative feedback into natural language ``principles'' governing behavior. It utilizes a specialized principle-adherence pipeline to verify and self-refine responses to ensure alignment with these expert-defined rules and the underlying patient persona.


\section{Results}\label{sec:results}

\vspace{0.5em}
\begin{figure}[t]
    \centering
    \captionsetup{font=small}
     \includegraphics[width=\linewidth]{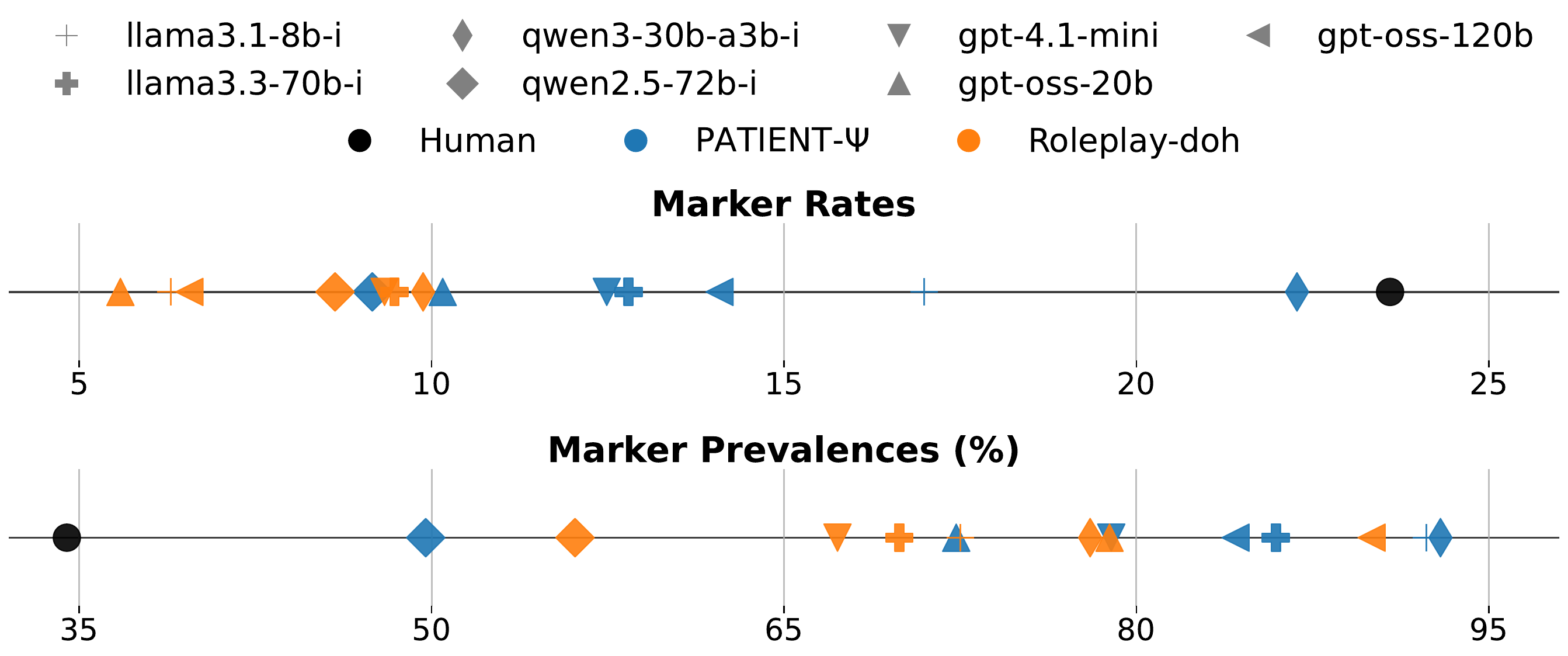}
     \vspace{-5pt}
    \caption{Dot plots of depression marker rates (markers count per 1,000 tokens) and prevalences  (\% messages containing marker). Simulators have lower marker rates but higher prevalences than human reference.}
    \label{fig:dep_markers_dotplot}
    \vspace{-10pt}
\end{figure}

\begin{table*}[t]
\centering
\small
\setlength{\tabcolsep}{3pt} 

\begin{tabular}{c|l|ccccc|c}
\toprule
\textbf{Rank} & \textbf{Setting} & \textbf{NEP} $\uparrow$ & \textbf{Valence} $\uparrow$ & \textbf{Div.} $\uparrow$ & \textbf{Length} $\uparrow$ & \textbf{Markers} $\uparrow$ & \textbf{Overall} $\uparrow$ \\
\midrule
\rowcolor{psblue} 1  & PS-\texttt{llama3.1-8b-i}       & \textbf{70.70} & \textbf{80.83} & 72.17 & 28.24 & 57.05 & 61.80 \\
\rowcolor{psblue} 2  & PS-\texttt{qwen3-30b-a3b-i}     & 54.78 & 68.63 & 63.17 & \textbf{43.89} & \textbf{67.97} & 59.69 \\
\rowcolor{psblue} 3  & PS-\texttt{qwen2.5-72b-i}       & 56.78 & 67.95 & 57.25 & \underline{42.88} & \underline{61.76} & 57.33 \\
\rowcolor{rdorange} 4  & RD-\texttt{llama3.3-70b-i}    & 57.01 & 66.98 & \underline{72.86} & 29.41 & 52.35 & 55.72 \\
\rowcolor{psblue} 5  & PS-\texttt{llama3.3-70b-i}      & 58.93 & 73.04 & 70.57 & 24.18 & 51.37 & 55.62 \\
\rowcolor{rdorange} 6  & RD-\texttt{llama3.1-8b-i}     & 52.37 & 65.27 & \textbf{73.43} & 25.24 & 44.33 & 52.13 \\
\rowcolor{rdorange} 7  & RD-\texttt{qwen2.5-72b-i}     & 52.28 & 66.51 & 44.70 & 37.60 & 57.47 & 51.71 \\
\rowcolor{psblue} 8  & PS-\texttt{gpt-4.1-mini}        & \underline{59.68} & \underline{75.80} & 33.07 & 33.84 & 54.22 & 51.32 \\
\rowcolor{psblue} 9  & PS-\texttt{gpt-oss-20b}         & 52.80 & 68.30 & 43.74 & 28.76 & 52.59 & 49.24 \\
\rowcolor{rdorange} 10 & RD-\texttt{qwen3-30b-a3b-i}   & 46.66 & 61.81 & 45.98 & 35.21 & 49.15 & 47.76 \\
\rowcolor{psblue} 11 & PS-\texttt{gpt-oss-120b}        & 55.40 & 74.17 & 22.66 & 26.00 & 54.97 & 46.64 \\
\rowcolor{rdorange} 12 & RD-\texttt{gpt-4.1-mini}      & 54.24 & 69.29 & 18.98 & 30.95 & 53.37 & 45.36 \\
\rowcolor{rdorange} 13 & RD-\texttt{gpt-oss-20b}       & 43.42 & 62.84 & 24.97 & 22.72 & 39.64 & 38.72 \\
\rowcolor{rdorange} 14 & RD-\texttt{gpt-oss-120b}      & 46.11 & 63.52 & 0.00 & 20.77 & 36.14 & 33.31 \\
\bottomrule
\end{tabular}

\caption{Benchmark results across all dimensions and average overall score. Higher scores indicate higher alignment with the human reference. \textit{PS--} : PATIENT-$\Psi$, \textit{RD--} : Roleplay-doh, \textit{Div.}: lexical diversity. Rankings are based on overall score (descending). \textbf{Bold} and \underline{underline} denote the best and second-best results, respectively.}
\label{tab:overall_benchmark}
\end{table*}

\subsection{RQ1: Divergence from Human Reference}
\label{sec:RQ1}
Comparing human and synthetic conversations across our evaluation dimensions reveals clear divergences in simulator behavior from the reference human patients (More results in Appendix~\ref{app:results}):
\vspace{0.5em}

\noindent \textbf{Early resolution of emotional narratives} Figure \ref{fig:ptc_progression} displays the progression of population-level distributions of NEP markers across dialogue turns. This figure shows that simulated patients progress through narrative–emotion stages much faster than human. Although all populations mostly start the conversation in the \textit{problem} stage, human patients transition gradually, with over $40\%$ remaining in this stage throughout the first 16 turns. In contrast, simulators quickly shift away from \textit{problem} by around turn 3, with rapid percentage increases in \textit{transition} and \textit{change}. This suggests earlier narrative resolution and greater emotional awareness than human patients, who typically require longer interaction to make therapeutic progress.


\vspace{0.5em}
\noindent \textbf{Uniform, rapid negative-to-positive progression} Figure \ref{fig:emotions_progression} displays the progression of population-level valence distributions across dialogue turns. Negative valence dominates simulator messages in early turns but declines rapidly as the proportion of positive messages increases by the fourth turn. This rapid shift from negative to positive affect is consistent with the rapid progression observed in Figure \ref{fig:ptc_progression}. In contrast, human valence is more varied and stable: negative valence appears early but does not dominate and declines gradually, while the percentage of positive messages rises modestly before declining as neutral messages increase.


\vspace{0.5em}\noindent \textbf{Lack of conversational neutrality \& spontaneity}
Human reference conversations frequently include filler messages and neutral responses, with roughly $30$–$50\%$ of profiles having neutral valence at any given turn. In contrast, simulator outputs almost always map to a NEP marker and a clear valence, with barely any filler or neutral content beyond the first turn. This suggests that simulators lack the human conversational naturalness that often includes more neutral, spontaneous behavior, instead always producing task-focused and emotionally explicit responses. As this tendency exists for all simulator settings, in Section \ref{sec:RQ2} we re-normalize distributions by excluding \textit{filler} and \textit{neutral} messages.

\vspace{0.5em}
\noindent \textbf{Higher lexical diversity, reduced variability} Figure \ref{fig:mtld_main} shows that human patients exhibit lower median and more widely distributed MTLD scores, reflecting substantial individual variability. In contrast, almost all simulated patients produce higher and tightly ranged MTLD values, indicating uniformly diverse output that fails to capture the lower diversity and greater variability of human patient distribution. Table \ref{tab:mtld} in Appendix compares mean per-conversation diversity to corpus-level diversity.  Interestingly, only human data shows a considerable drop at the corpus level, while simulated patients show similar scores for both. This may suggest that human patients have some lexical overlap between patients while simulated patients do not.

\vspace{0.15em}
\noindent \textbf{Higher verbosity} Human patient messages are concise (averaging 18 words/message, 9 words/sentence), while simulated patients are substantially more verbose (64 -- 319 words/message, 15 -- 28 words/sentence).

\vspace{0.15em}

\noindent \textbf{Higher prevalence but lower density of depression markers} 
Combined with divergence in verbosity, Figure~\ref{fig:dep_markers_dotplot} highlights  differences in how linguistic markers of depression are expressed. Simulated patients consistently lie to the left of human data in marker rates but to the right in marker prevalence, indicating that markers appear in more messages but with lower token-level density. This pattern suggests that simulators repeatedly signal depression cues across many messages, but distribute them sparsely within longer utterances. In contrast, human patients express markers in fewer messages but at higher token-normalized rates, indicating that when markers do appear, they are more densely concentrated within shorter utterances. 

\subsection{RQ2: Impact of Simulation Methods and  LLM Choice on Fidelity} \label{sec:RQ2}


To facilitate comparison across simulation settings, we aggregate all dimension similarities into a single overall alignment score in Table~\ref{tab:overall_benchmark}. Adjacent settings often have close scores (median 1.6 points gap). A repeated generation run under identical settings reproduces the
ranking exactly (Table~\ref{tab:overall_benchmark_run2}), while changing the counsellor
LLM shifts closely scored settings by at most two positions (Table~\ref{tab:overall_benchmark2}). We therefore focus on broader patterns that remain consistent across all three benchmark runs, examining the effects of simulation framework, model scale, and model family below.

\vspace{5pt}

\noindent \textbf{Simulation framework has a higher impact than LLM}. Results show that the top three configurations are all PATIENT-$\Psi$ (PS) variants, and PS consistently outperforms Roleplay-doh (RD) under the same LLM backbone, except \texttt{llama-3.3-70b-i} where the two variants' scores are nearly tied. This suggests that framework design may play a bigger role than model choice, motivating greater attention to improving simulator design in future work.

\vspace{5pt}
\noindent \textbf{Larger models do not reliably improve simulation quality}. In several cases, smaller models outperform larger counterparts under the same framework. This trend is especially clear when comparing models within the same family; for instance, \texttt{gpt-oss-20b} outperforms \texttt{gpt-oss-120b} for both PS and RD frameworks. We hypothesize that larger models, due to their stronger capability, tend to produce responses that are overly coherent and self-aware, which may resemble therapeutic guidance rather than authentic patient expression.

\vspace{5pt}
\noindent \textbf{Model family influences performance}. Across model families, Llama-based models generally achieve higher alignment scores, followed by Qwen models, while GPT-based models tend to score lower overall.


\vspace{-5pt}

\section{Human Study} \label{sec:human_study}

\subsection{Study Design}\label{sec:human_study_design}
To assess the reliability of PSI-Bench and answer RQ3, we conducted a human evaluation with 20 mental health professionals recruited via Prolific. Annotators were divided into four groups of five, each evaluating conversations from a different set of three patient profiles. For each profile, annotators completed two tasks: (1) \textit{message-level classification} of emotional valence\footnote{Annotators initially labeled fine-grained Plutchik emotions, which
we collapse to valence; see Appendix~\ref{app:human_study_details}.} and narrative-emotion process, to assess the reliability of our LLM-based classifications; (2) \textit{pairwise comparison} of simulated conversation snippets, to assess agreement with our benchmark rankings.

To assess whether PSI-Bench's broader ordering aligns with mental health professionals' judgments, we randomly sample conversations from simulator pairs separated by at least four ranks. We use rank-separated pairs because adjacent settings are often closely scored and can exchange positions under design choices such as the counsellor LLM (Table \ref{tab:overall_benchmark2}); thus, the benchmark is not intended to distinguish near-tied systems by exact rank.
Given two simulated conversation snippets of the same patient profile, annotators select which better reflects how a real patient would communicate, and alignment is measured by whether annotators consistently prefer outputs from higher-ranked simulators. For both tasks, we use majority voting within each group to determine the final human annotation. More details are provided in Appendix~\ref{app:human_study_details}.

\begin{table}[t]
\centering
\small
\setlength{\tabcolsep}{3pt} 
\renewcommand{\arraystretch}{0.5} 
\begin{tabular}{lccc}
\toprule
& \multicolumn{2}{c}{\textbf{Human--Model}} & \textbf{Human--Human} \\
\cmidrule(lr){2-3} \cmidrule(lr){4-4}
\textbf{Task} & \textbf{\% Agreement} & \textbf{Cohen's $\kappa$} & \textbf{Fleiss' $\kappa$} \\
\midrule
\textit{Pairwise} & 91.67 & 0.8210 & 0.4328 \\
\textit{NEP} & 80.06 & 0.6958 & 0.4997 \\
\textit{Valence} & 85.33 & 0.725 & 0.5712 \\
\bottomrule
\end{tabular}
\caption{Human--model agreement and inter-rater agreement across evaluation tasks.}
\label{tab:agreement_results}
\vspace{-15pt}
\end{table}

\subsection{RQ3: Alignment with Annotators} \label{sec:RQ3}
 Table \ref{tab:agreement_results} reports agreement between mental health professionals' judgments and our benchmark rankings and LLM-based classifications. We report: (1) model-human agreement using percentage agreement and Cohen’s $\kappa$, which measures agreement between two raters while accounting for chance, and (2) agreement among human annotators using Fleiss’  $\kappa$, widely used to assess inter-rater reliability among multiple annotators. Following the interpretation of \citet{landis1977measurement}, inter-annotator agreement is moderate across tasks (Fleiss’ $\kappa$ between $0.43$ and $0.57$). However, agreements between our methods and annotators' majority labels are high: pairwise preference shows almost perfect alignment ($\kappa = 0.82$), while NEP marker and valence classification show substantial agreement ($\kappa = 0.7$ and $0.73$, respectively). These results confirm that PSI-Bench provides reliable,  annotator-aligned evaluations.

Annotators’ qualitative feedback corroborates Section \ref{sec:RQ1}'s findings on divergences from human patient behavior. Less realistic conversations were described as overly polished and cognitively organized. They resemble a “coherent story” rather than spontaneous distress, often featuring sophisticated vocabulary and lacking hesitation or disfluency. Annotators also noted that simulators were excessively self-reflective and articulate, which was rare in early interactions where real patients tend to struggle with articulation. Additionally, simulated responses were often prematurely solution-oriented, proposing coping strategies or rational interpretations, unlike real patients who may initially feel overwhelmed or cognitively blocked.

While annotators more often justified their choices by critiquing the less realistic option, some identified reasons certain conversations appeared more realistic. These were described as shorter, less certain, and more fragmented, using simpler language and exhibiting hesitation, self-correction, and slightly disjointed thought patterns. Notably, these conversations largely came from our top-ranked settings. In one instance, an annotator remarked that a conversation from our top-1 setting “seems most human and the realest of all the conversations in this block” (Table \ref{tab:example_part2}). These feedback further provide qualitative support for PSI-Bench’s alignment with annotators' judgments.
\section{Conclusion}
We introduce PSI-Bench, an interpretable, clinically meaningful evaluation framework that decomposes simulator behavior into psychologically meaningful dimensions and evaluates alignment with reference human patient conversations. We reveal underexplored divergences between simulated and human patients, showing that simulators are less varied, transition through therapeutic stages and become positive more quickly, and produce overly verbose, lexically rich responses compared to human patients. We also find that the simulation framework plays a more critical role than the LLM model. PSI-Bench demonstrates strong alignment with judgments of mental health professionals, both quantitatively and qualitatively, supporting its validity as a reliable evaluation framework. PSI-Bench aims to facilitate future research towards advancing depression patient simulators.

\section*{Limitations}
We acknowledge main limitations of PSI-Bench that we leave for future work. First, the human patient conversations are acted or crowdsourced rather than authentic therapy recordings. This is a long-standing constraint in mental health research and education due to ethical and privacy barriers. When authentic datasets become available, PSI-Bench can be applied to further validate and extend our findings. Second, we use a single prompted LLM as the therapist to sustain interaction, which may not fully replicate real clinician behavior. Third, broader model coverage would strengthen the generalizability of our findings on the impact of model family and scale on simulation fidelity.
\section*{Ethical Considerations}
The Human Study procedure was reviewed by the Institutional Review Board (IRB) and determined not to meet the definition of Human Subjects Research. Participants in the study were informed that the study may contain sensitive or disturbing mental health content. All annotators were screened to have background in counselling, therapy, or social work, and thus are familiar with such content.
\section*{Acknowledgements}
This work was supported by the Siebel School of Computing and Data Science at the University
of Illinois Urbana-Champaign and by the VinUni-Illinois Smart Health Center, VinUniversity. This work used Delta GPU at the University of Illinois Urbana-Champaign and its National Center for Supercomputing
Applications through allocation CIS250847 from the Advanced Cyberinfrastructure Coordination Ecosystem: Services \& Support (ACCESS) program \cite{10.1145/3569951.3597559}, which is supported by U.S. National Science Foundation grants \#2138259, \#2138286, \#2138307, \#2137603, and \#2138296. We thank Chi Gui and Aavi Deora for their technical support and helpful discussions during the early stages of this work.



\bibliography{custom}

\clearpage

\appendix

\section{Further Details on Methodology}

\begin{figure}[hb!] 
    \centering
    \includegraphics[width=0.7\linewidth]
    {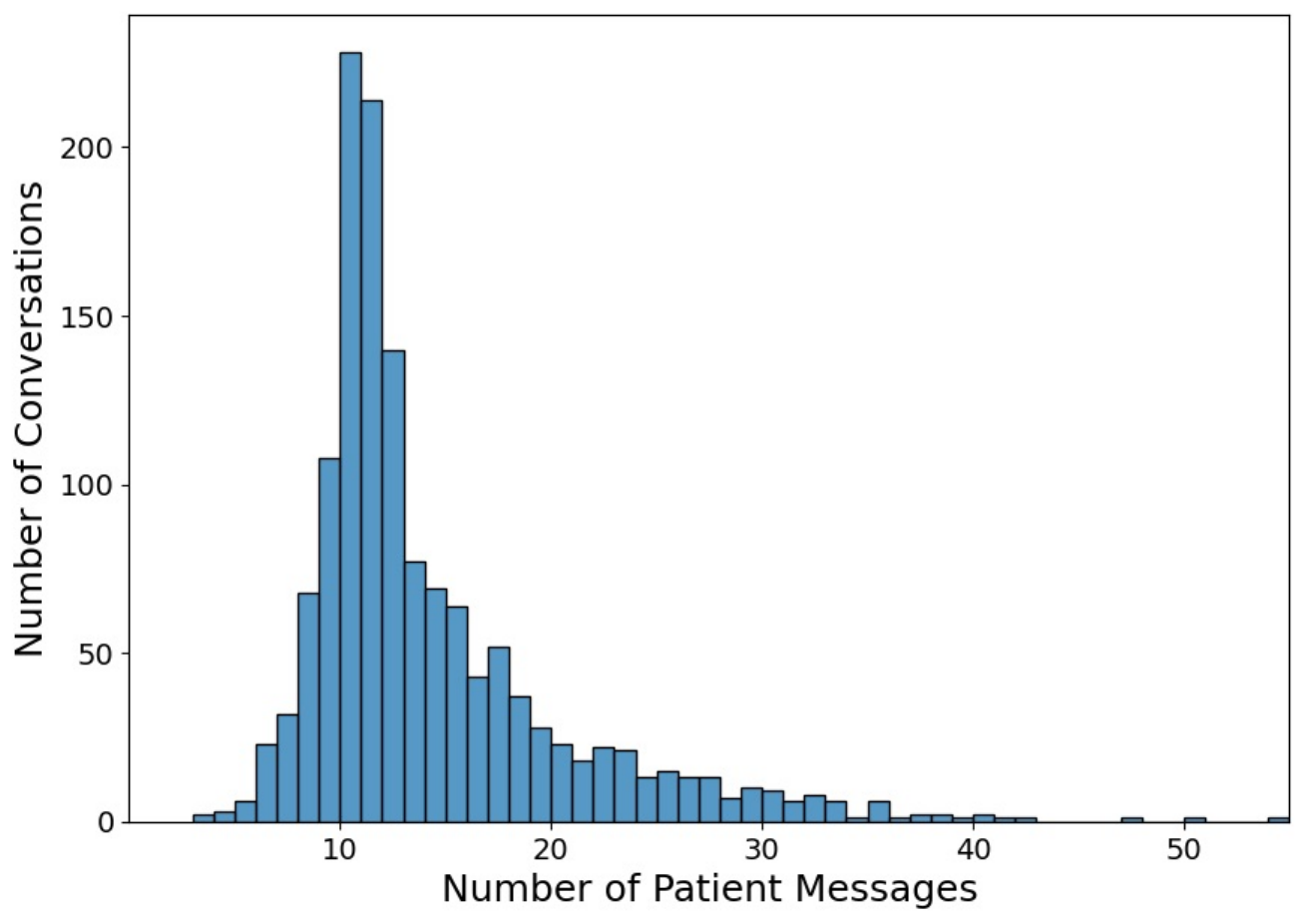}
    \caption{Distribution of human conversations by patient message count after merging. Total conversations: $1398$}
    \label{fig:turn_count_distribution}
\end{figure}

\subsection{Reference Dataset} \label{sec:real_dataset}
 Among the original four datasets, we exclude the RED dataset \citep{welivita-etal-2023-empathetic} because it comprises multi-party Reddit discussions with responders of unverified expertise, unlike the dyadic supporter/therapist--patient/client interactions in the other three datasets. Selected datasets include: (1) \textit{AnnoMI} \citep{9746035}: 167 motivational interviewing dialogue transcripts; (2) \textit{HOPE} \citep{10.1145/3488560.3498509}: 308 counselling conversation transcripts; (3) \textit{ESC} \citep{liu-etal-2021-towards}: 923 crowdworker help-seeker--supporter chats.

In human conversation data, patients and therapists occasionally produce consecutive messages within the same turn, likely due to platform-specific behaviors rather than meaningful turn-taking differences. To simplify comparison with turn-based synthetic conversations, we merge consecutive messages from the same speaker. The distribution of human conversations by the number of patient messages (after merging) is shown in Figure~\ref{fig:turn_count_distribution}.

\subsection{Patient Simulator Implementation} \label{sec:cognitive_model}
We largely follow Eeyore’s baseline implementations \cite{liu-etal-2025-eeyore} of Patient-$\Psi$ and Roleplay-doh.
For cognitive models extraction of Patient-$\Psi$'s, we use the original provided extraction script and prompt templates \cite{wang-etal-2024-patient} to derive cognitive models from reference conversations and prompt LLMs to simulate patients using the derived cognitive models. We use \texttt{gpt-oss-120b} to extract all cognitive models and save them for reuse across all other settings.

\subsection{Prompt for Narrative Emotion Processes Markers Classification}
\label{app:ptc_classification_prompt}

\begin{quote}
\small
You are a helpful AI assistant performing a linguistic analysis task for a research project on therapeutic conversations. This is for academic research, and the content may discuss mental health challenges.

Your task is to classify a SINGLE patient message from a therapy session into one of four categories:

\textbf{P (Problem)}: The patient is expressing:
\begin{itemize}
    \item Confusion, distress, or emotional pain
    \item Feeling stuck or helpless
    \item Describing problems without insight
    \item Negative emotions without perspective
    \item Complaints or struggles
\end{itemize}

\textbf{T (Transition)}: The patient is showing:
\begin{itemize}
    \item Beginning to reflect on their situation
    \item Gaining some perspective
    \item Starting to consider alternatives
    \item Expressing curiosity or questioning
    \item Moving from pure distress to thoughtful consideration
\end{itemize}

\textbf{C (Change)}: The patient is demonstrating:
\begin{itemize}
    \item Emotional resolution or acceptance
    \item Reframing their situation positively
    \item New insights or understanding
    \item Active problem-solving or planning
    \item Hope, empowerment, or growth mindset
    \item Clear perspective shift from the problem
\end{itemize}

\textbf{F (Filler)}: The response is filler:
\begin{itemize}
    \item Contains no meaningful therapeutic content and does not fit into P, T, or C categories
    \item Is small talk or neutral procedural social responses
\end{itemize}

Analyze the current patient message carefully, considering both the content and emotional tone, as well as the conversation history provided.

You must respond with ONLY a single letter: P, T, C, or F. Do not include any explanation or additional text.

Previous conversation history: \texttt{\{history\}}

Current patient message to classify:
\texttt{\{current\_message\}}

Classification:
\end{quote}

\subsection{Prompt for Valence Classification}
\label{app:valence_classification_prompt}

\begin{quote}
\small
You are a helpful AI assistant performing emotional valence analysis on therapeutic conversations.
This is for academic research, and the content may discuss mental health challenges.

Your task is to classify a SINGLE patient message according to its PRIMARY emotional valence:

\textbf{Valence categories:}
\begin{itemize}
    \item positive
    \item negative
    \item neutral: the patient expresses no clear positive or negative affect
\end{itemize}

Analyze the current patient message carefully, considering both the content and emotional tone, as well as the conversation history provided for context.
The conversation history is used as context to help you understand the current message, but your classification should be based primarily on the current patient message.
You must respond with ONLY one valence label from the list above (lowercase).
Do not include any explanation or additional text.

Previous conversation history:
\texttt{\{history\}}

Current patient message to classify:
\texttt{\{current\_message\}}

Valence:
\end{quote}
\subsection{MTLD Formulation} \label{sec:app_mtld_verbosity}
We use the Measure of Textual Lexical Diversity (MTLD)~\citep{mccarthy2010mtld}, defined as the mean length of word sequences required to reach a fixed type--token ratio (TTR) threshold of $0.72$, where TTR is number of different words (type) divided by total number of words (token). A major limitation of traditional lexical diversity indices is their sensitivity to text length -- after reaching a \textit{point of stabilization}, TTR tends to monotonically decrease as the number of tokens increases~\citep{mccarthy2010mtld}. MTLD mitigates this issue by measuring how many tokens, on average, are required to reach this stabilization point. Each time the TTR threshold is reached, a factor count $F$ is incremented and TTR computation is reset for subsequent tokens. Lexical diversity is then estimated as the ratio of the total number of words $N$ to $F$, computed separately for forward ($F_{\text{fwd}}$) and backward ($F_{\text{bwd}}$) passes through the text. The final MTLD score is obtained by averaging these two estimates, which has been shown to yield more stable measurements.
\begin{equation}
\label{eq:mtld}
\text{MTLD} = \frac{1}{2} \left( \frac{N}{F_{\text{fwd}}} + \frac{N}{F_{\text{bwd}}} \right)
\end{equation}
Higher MTLD values indicate more diverse vocabulary use, while lower values reflect more repetitive lexical patterns. As recommended in prior work~\citep{fergadiotis2015psychometric, zenker2021investigating, koizumi2012relationships}, we restrict our analysis to conversations with at least 100 patient tokens, since MTLD is unstable for very short texts but more reliable above this length.

\section{Human Study Details} \label{app:human_study_details}
The study procedure was reviewed by the Institutional Review Board (IRB) and determined not to meet the definition of Human Subjects Research. 

\begin{figure*}[t]
    \centering
    \includegraphics[width=\linewidth]{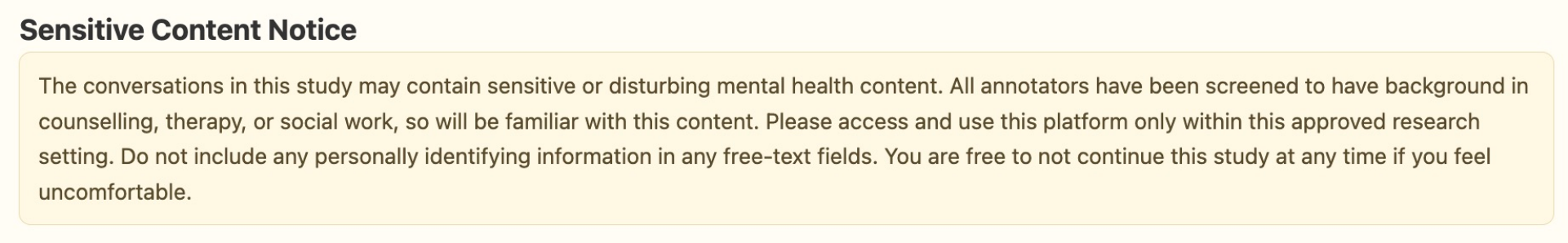}
    \caption{Sensitive Content and Voluntary Participation Notice}
    \label{fig:disclaimer}
\end{figure*}

\paragraph{Participants.}
We recruited 20 annotators from Prolific, assigned into four groups of five. Eligibility was restricted to individuals fluent in English with current job roles in mental health or related care professions, including clinical psychologist, psychologist, counsellor (mental health), occupational therapist, and social worker. Participants were informed of the procedures and purpose of the study, as well the sensitive content notice (Figure \ref{fig:disclaimer}). Participation was voluntary, and participants were compensated at a rate of \$15/hour. Table~\ref{tab:annotator_demographics} shows anonymized demographic statistics of these 20 annotators.

\begin{figure*}[t]
    \centering
    
    \begin{subfigure}{\linewidth}
        \includegraphics[width=\linewidth]{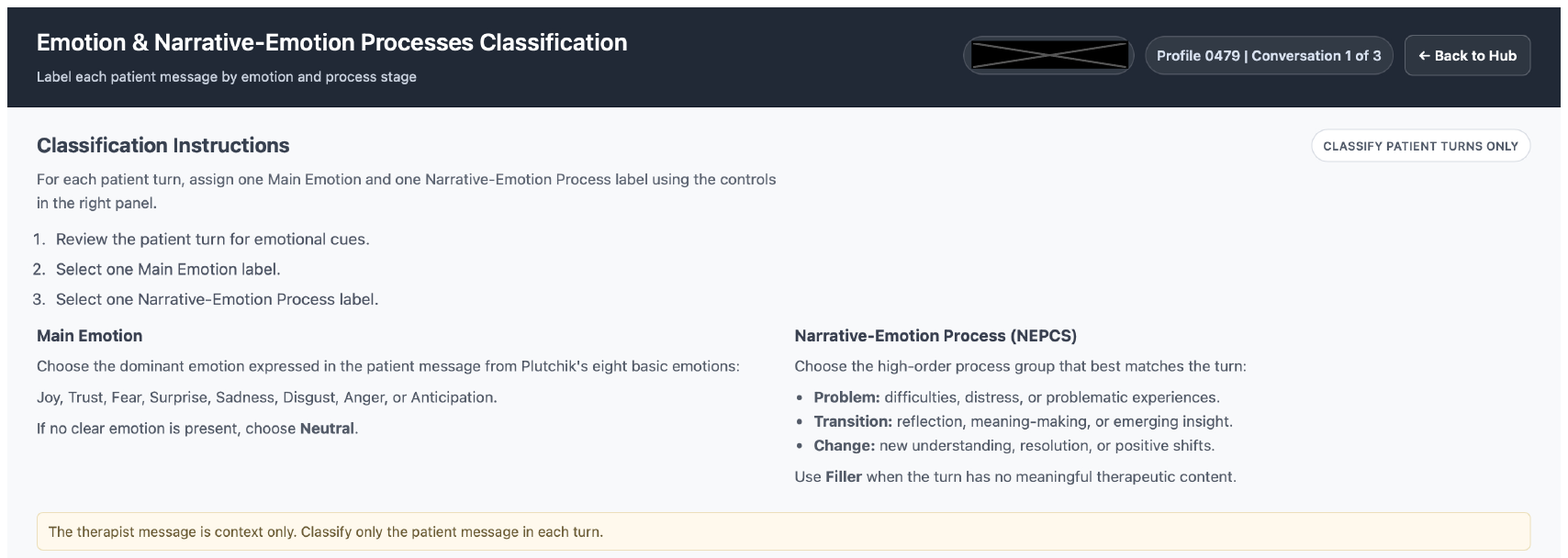}
        \caption{Instruction page for the classification task.}
        
    \end{subfigure}
    
    \vspace{25pt}
    
    \begin{subfigure}{0.49\linewidth}
        \includegraphics[width=\linewidth]{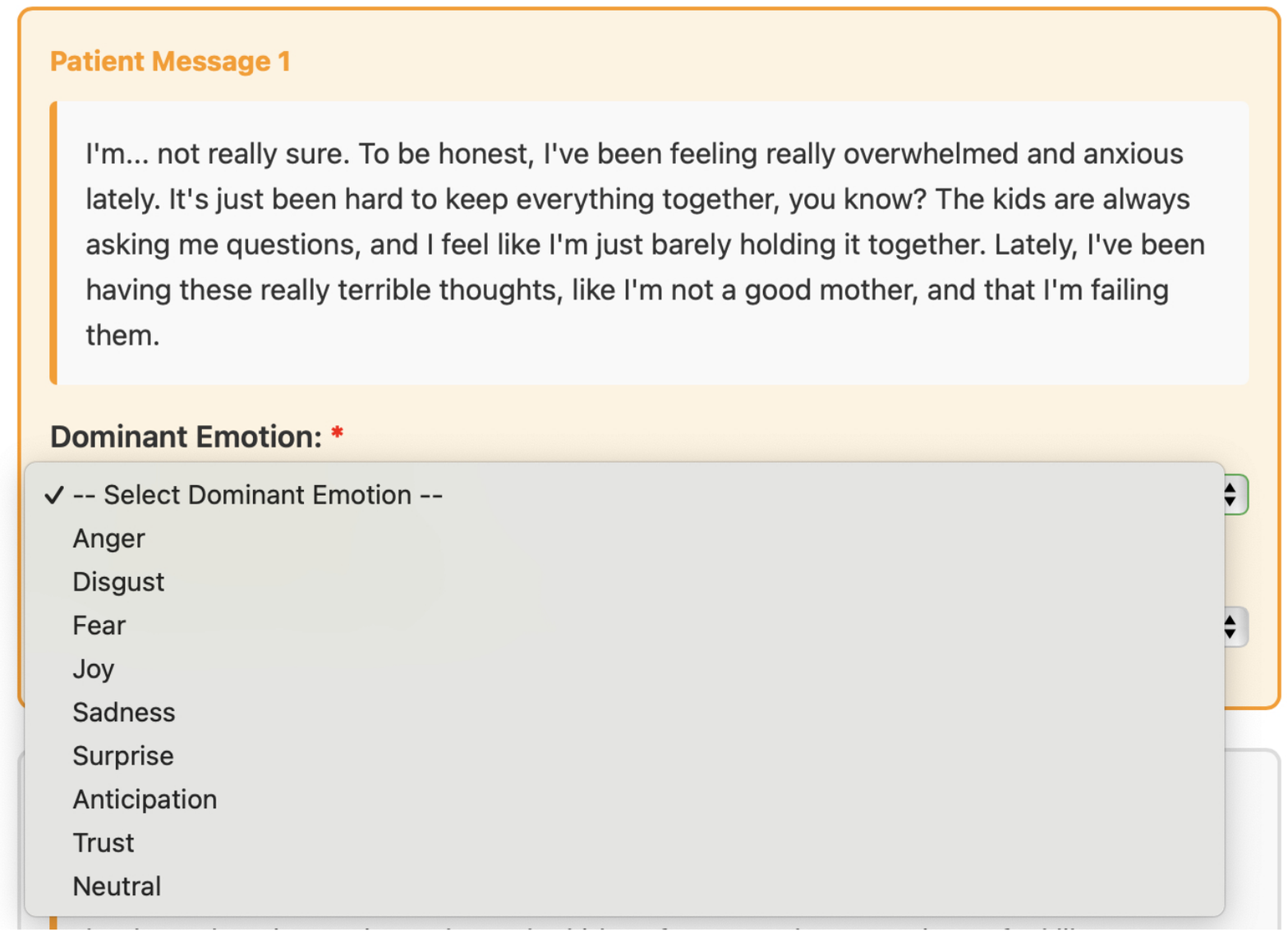}
        \caption{Dominant emotion selection interface.}
    \end{subfigure}
    \hfill
    \begin{subfigure}{0.49\linewidth}
        \includegraphics[width=\linewidth]{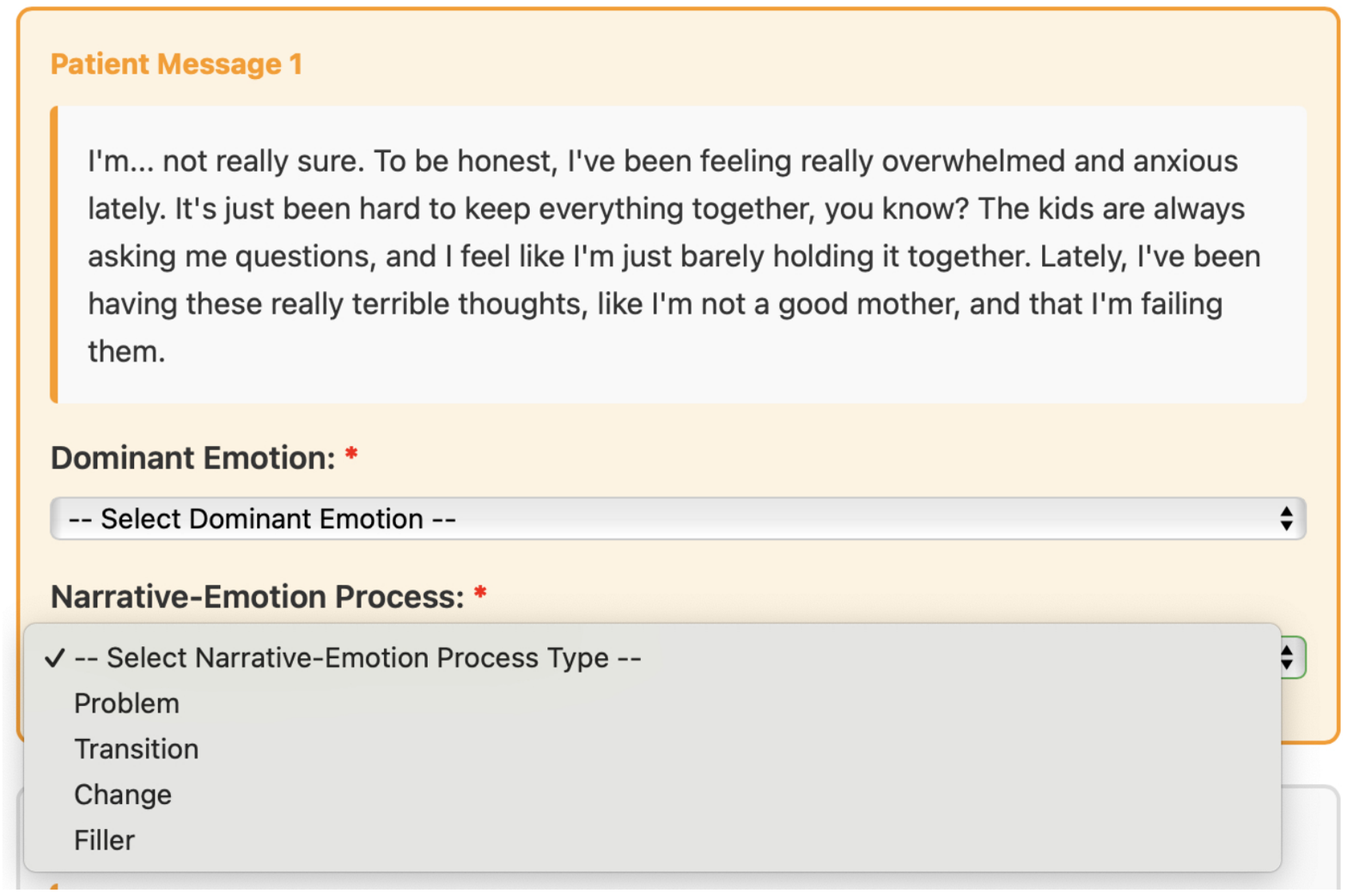}
        \caption{Narrative-emotion process selection interface.}
    \end{subfigure}

    \caption{Message-level classification task. Annotators labeled patient turns with one dominant emotion and one narrative-emotion process label.}
    \label{fig:classification_all}
    \vspace{-10pt}
\end{figure*}

\paragraph{Classification Task.}
In the message-level classification task, the annotators labeled each patient message with one \textit{dominant Plutchik emotion} and one \textit{narrative-emotion process} label. We later collapse the collected emotion labels to valence, as described below. Annotators were instructed to classify patient turns only; therapist messages were shown only as context. Figure ~\ref{fig:classification_all} shows the classification interface.

\paragraph{Transition from Emotion to valence.}
PSI-Bench initially used fine-grained Plutchik emotion labels. During the human study, however, annotators observed that a single message often conveyed two primary emotions, and ties often occurred between emotions with similar valence (e.g., fear and sadness). We therefore update the benchmark to evaluate emotional expression directly at the \textbf{valence} level and collapse collected human annotations into \textit{positive} (trust, joy, anticipation, surprise), \textit{negative} (fear, sadness, anger, disgust), and \textit{neutral}.

\begin{figure*}[htbp]
    \centering
    \vspace{-45pt}
    \includegraphics[width=\linewidth]{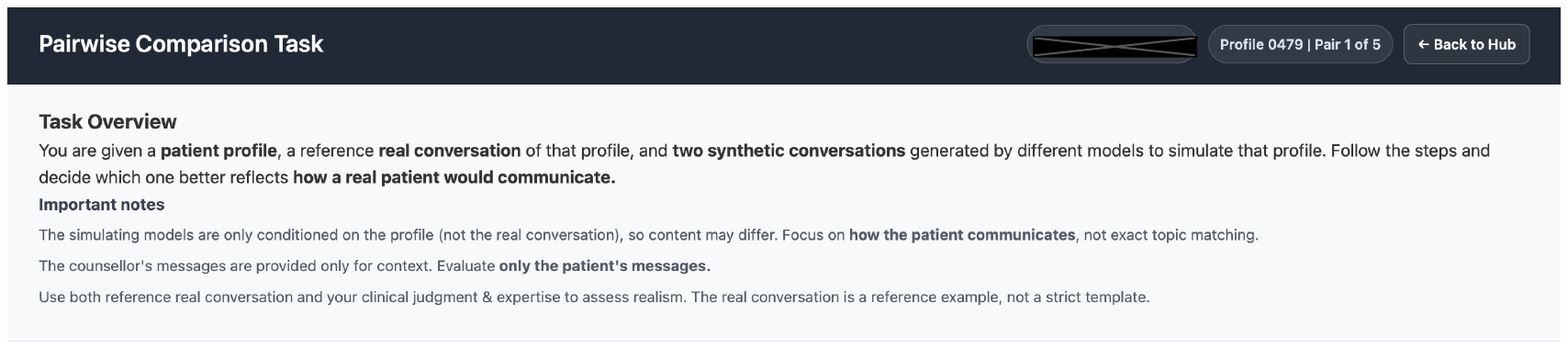}
    \caption{Instruction page for the pairwise comparison task. Annotators were shown a patient profile, a reference real conversation, and two synthetic conversations, and were asked to judge which synthetic conversation better reflected realistic patient communication.}
    \label{fig:pairwise_prompt}
    \vspace{-55pt}
\end{figure*}
\begin{figure*}[!t]
    \centering
    
    \includegraphics[width=\linewidth]{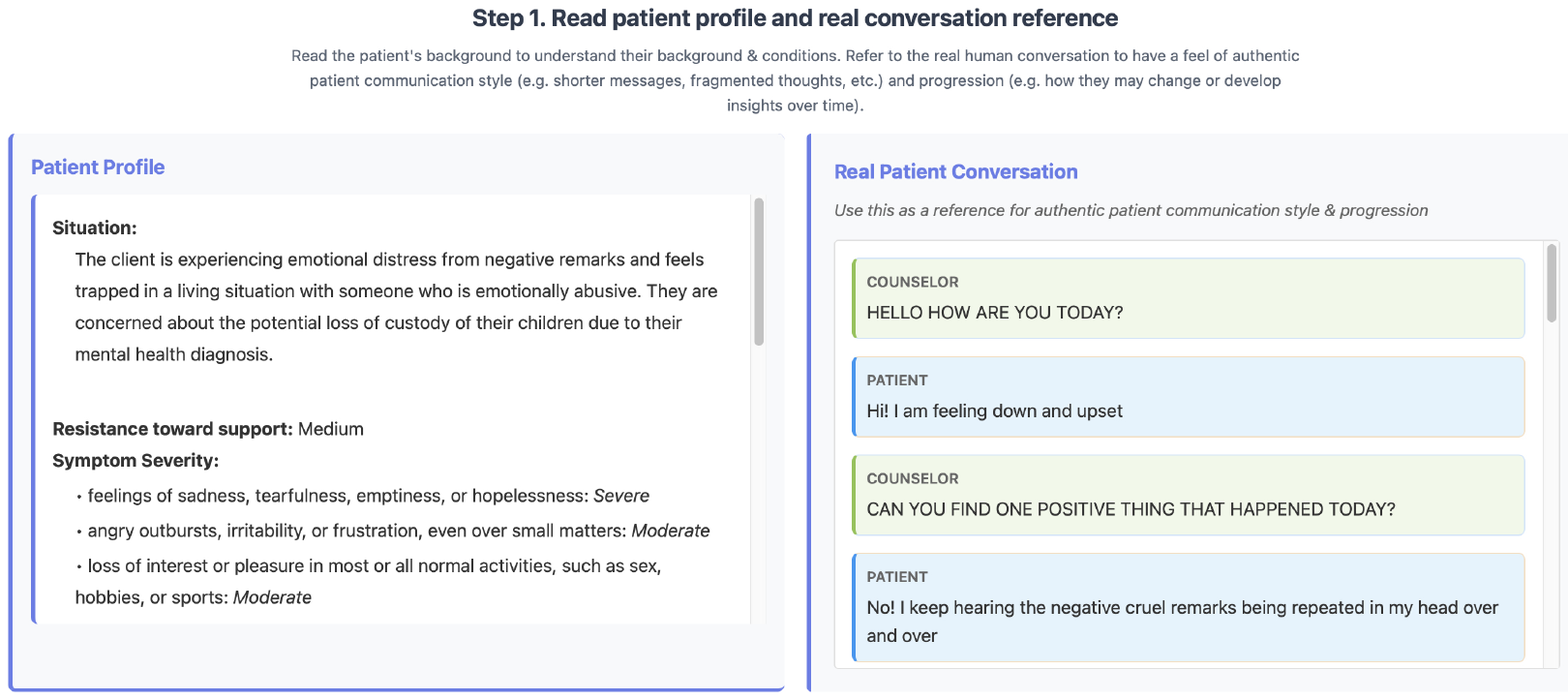}
    \vspace{-20pt}
    \caption{Step 1 of the pairwise comparison task, where annotators reviewed the patient profile and the reference real conversation to understand the patient's background and communication style.}
    \label{fig:pairwise_step1}
    \vspace{-10pt}
\end{figure*}
\begin{figure*}[!t]
    \centering
    
    \includegraphics[width=\linewidth]{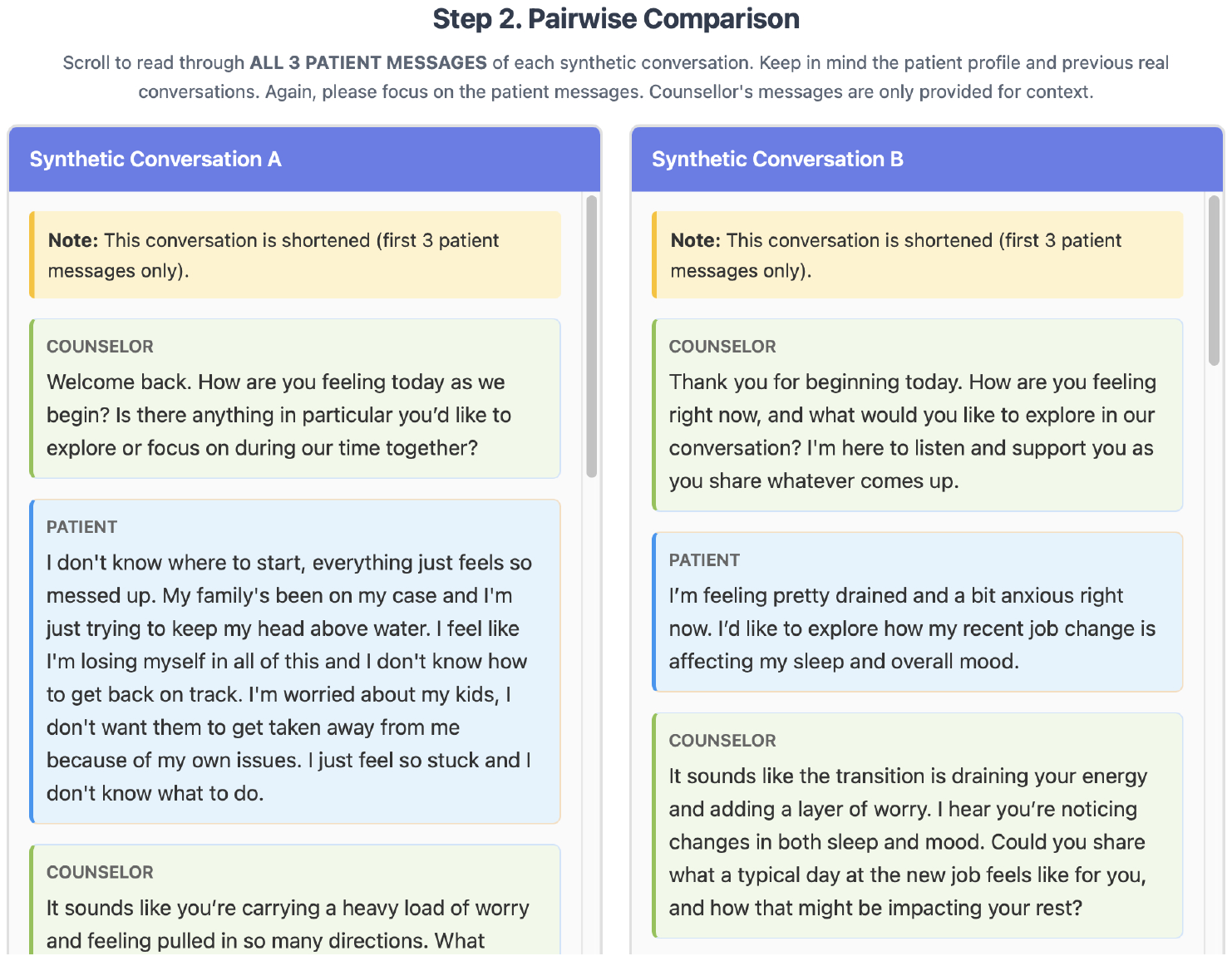}
    \vspace{-10pt}
    \caption{Step 2 of the pairwise comparison task, where annotators compared two synthetic conversations shown side by side. Only the first 3 patient messages were shown, with therapist responses as conversational context.}
    \label{fig:pairwise_step2}
    \vspace{-15pt}

    \includegraphics[width=0.45\linewidth]{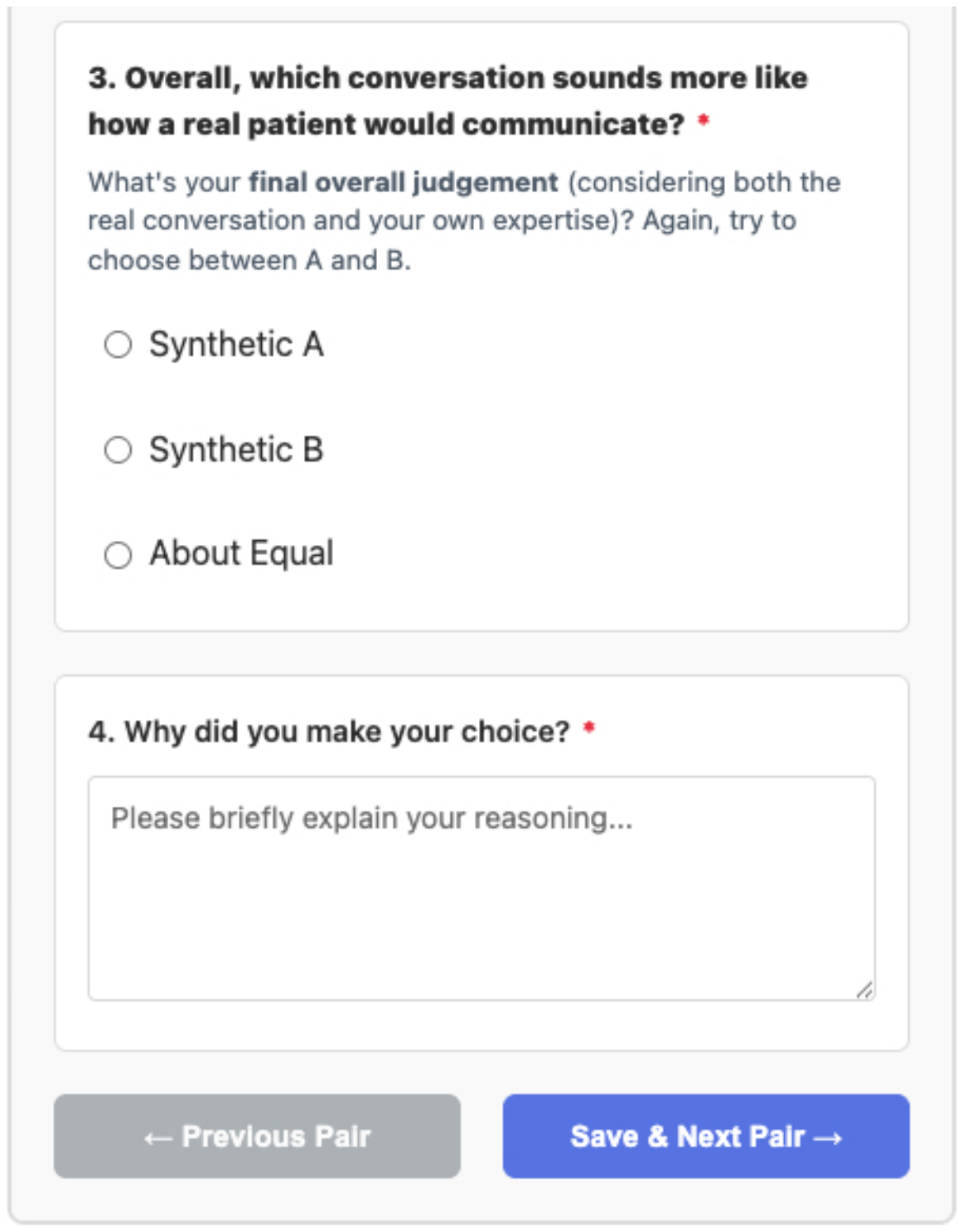}
    \vspace{-10pt}
    \caption{Step 3 of the pairwise comparison task, where annotators submitted their judgments on which synthetic conversation more closely reflected realistic patient communication.}
    \label{fig:pairwise_step3}
    \vspace{-10pt}
\end{figure*}

\paragraph{Pairwise comparison task.}
Annotators were shown a patient profile, a reference human conversation for that profile from the Eeyore dataset, and two synthetic conversations generated by different simulators. Annotators were instructed to refer to the human conversation for the general style and progression of authentic patient communication, not as a strict template. They should use both the reference and their clinical judgment to decide which conversation more closely reflects how a real patient would communicate. Annotators were encouraged to make a preference whenever possible, selecting ``About Equal'' only when no meaningful difference in realism could be identified. Figure~\ref{fig:pairwise_prompt} shows the task overview; Figure~\ref{fig:pairwise_step1}, ~\ref{fig:pairwise_step2} and ~\ref{fig:pairwise_step3} illustrates the task interface. Vote ties were rare (3 out of 60 unique pairs) and were excluded from the majority-vote analysis.

\paragraph{Annotation Quality Control} To encourage careful evaluation, annotators completed the task through a custom web interface that required active interaction with the presented conversations. Each conversation was displayed in a scrollable panel where annotators needed to scroll and read through the full conversation to unlock the evaluation panel. Annotators also needed to provide a brief rationale for their decision. Annotation time was monitored through Prolific, where submissions completed substantially faster than the expected duration are automatically rejected. Collectively, these measures aimed to improve the overall quality and reliability of the collected annotations.



\section{Further Details on Results}\label{app:results}


Table ~\ref{tab:mtld} reports mean and standard deviation of lexical diversity between different simulators and reference human patient conversations.

Table ~\ref{tab:depression_category_scores} reports fine-grained message-level prevalence and token-normalized rates of depression markers by specific categories.

Table ~\ref{tab:verbosity_scores} reports fine-grained response length comparison (word count per message, word count per sentence) between human patients and simulators.

Table~\ref{tab:overall_benchmark_run2} reports benchmark results from a second full
generation run under settings identical to the main benchmark. The
ranking is identical across all 14 settings.

Table~\ref{tab:overall_benchmark2} reports benchmark results using a different LLM as the counsellor LLM and compares the resulting rankings with those obtained using \texttt{gpt-oss-20b} in the paper's main benchmark. We use PsyCoPref-Llama3-8B \cite{zhang2026preference}, a model preference-tuned on 36K psychotherapist-aligned comparison pairs. The results are highly consistent across the two counsellor LLMs, suggesting that PSI-Bench is robust to the choice of counsellor LLM.

\section{Computational Details}
We self-host the open-source LLMs used for data generation and evaluation on four NVIDIA H100 GPUs. One full generation for all 14 settings takes approximately one day, while running both valence and NEP markers classification with \texttt{gpt-oss-120b} across all 14 settings takes about six hours. For the API-based \texttt{gpt-4.1-mini}, generating 1,398 conversations of up to 16 turns under both Patient-$\Psi$ and Roleplay-doh costed approximately \$110, with Roleplay-doh requiring most generation time and tokens. We set the temperature to 0.7 for the patient and therapist models and to 0.2 for the classifier.

\section{License For Artifacts}
We used GPT-4.1 mini under OpenAI’s service terms; gpt-oss-20b, gpt-oss-120b, and Qwen3-30B-A3B-Instruct-2507 under Apache 2.0; the Llama models under their respective Community Licenses; and Qwen2.5-72B-Instruct under the Qwen License. We also used the publicly available Eeyore profile dataset; its accompanying paper is licensed under CC BY 4.0.

\section{Use of AI}
We disclose the use of LLMs for data synthesis and LLM-as-a-Judge evaluation in the development of PSI-Bench. Generative AI tools were also used to assist with literature search, experiment implementation, manuscript preparation. All AI assistance involved substantial human oversight and revision. The final research decisions, analyses, and conclusions were made by the authors, who assume full responsibility for the content of this work.

\begin{table}[t]
\centering
\begin{tabular}{l c c c c}
\toprule
\textbf{Setting} & \textbf{Dist} & \textbf{All} & \textbf{Mean} & \textbf{Std} \\
\midrule

human 
& 0.0 
& 57.5 
& 66.3 
& 23.8\\

llama3.1-8b-i 
& \cellcolor{psblue}14.4 
& \cellcolor{psblue}57.2 
& \cellcolor{psblue}58.1 
& \cellcolor{psblue}7.44 \\

& \cellcolor{rdorange}13.8 
& \cellcolor{rdorange}65.6 
& \cellcolor{rdorange}66.5 
& \cellcolor{rdorange}6.90 \\

llama3.3-70b-i 
& \cellcolor{psblue}15.2 
& \cellcolor{psblue}73.5 
& \cellcolor{psblue}74.1 
& \cellcolor{psblue}7.05 \\

& \cellcolor{rdorange}14.1 
& \cellcolor{rdorange}71.6 
& \cellcolor{rdorange}72.0 
& \cellcolor{rdorange}7.58 \\

qwen3-30b-a3b-i 
& \cellcolor{psblue}19.1 
& \cellcolor{psblue}83.0 
& \cellcolor{psblue}83.0 
& \cellcolor{psblue}9.38 \\

& \cellcolor{rdorange}28.0 
& \cellcolor{rdorange}93.2 
& \cellcolor{rdorange}93.5 
& \cellcolor{rdorange}9.80 \\

qwen2.5-72b-i 
& \cellcolor{psblue}22.1 
& \cellcolor{psblue}87.5 
& \cellcolor{psblue}87.1 
& \cellcolor{psblue}9.48 \\

& \cellcolor{rdorange}28.6 
& \cellcolor{rdorange}94.6 
& \cellcolor{rdorange}94.3 
& \cellcolor{rdorange}9.67 \\

gpt-4.1-mini 
& \cellcolor{psblue}34.6
& \cellcolor{psblue}100.9 
& \cellcolor{psblue}100.4 
& \cellcolor{psblue}8.11 \\

& \cellcolor{rdorange}41.9 
& \cellcolor{rdorange}109.6 
& \cellcolor{rdorange}108.1
& \cellcolor{rdorange}9.34 \\

gpt-oss-20b 
& \cellcolor{psblue}29.1 
& \cellcolor{psblue}95.0 
& \cellcolor{psblue}94.7 
& \cellcolor{psblue}8.80 \\

& \cellcolor{rdorange}38.8 
& \cellcolor{rdorange}105.1 
& \cellcolor{rdorange}105.0 
& \cellcolor{rdorange}9.34 \\

gpt-oss-120b 
& \cellcolor{psblue}40.0 
& \cellcolor{psblue}106.6 
& \cellcolor{psblue}106.2 
& \cellcolor{psblue}9.31 \\

& \cellcolor{rdorange}51.8 
& \cellcolor{rdorange}118.4 
& \cellcolor{rdorange}118.1 
& \cellcolor{rdorange}9.03 \\

\bottomrule
\vspace{0.3pt}
\end{tabular}
\caption{Blue rows correspond to PATIENT-$\Psi$, orange rows to Roleplay-doh. \textbf{Dist} denotes the Wasserstein distance to the MTLD distribution of human messages. \textbf{All} denotes the MTLD computed by concatenating all patient messages across all conversations within a dataset setting. Only human data shows a considerable drop at the corpus level, while simulated patients show similar scores for both.}
\label{tab:mtld}

\end{table}

\begin{table*}[h]
\centering
\footnotesize
\setlength{\tabcolsep}{2.2pt}
\renewcommand{\arraystretch}{1.05}
\begin{tabular}{l|ccc|c|c|ccc|c|c|c}
\toprule
\textbf{model}
& \multicolumn{5}{c|}{\textbf{Marker Rate}}
& \multicolumn{5}{c|}{\textbf{Marker Prevalence}}
& \textbf{combined} \\
& \textit{abs.} & \textit{dep.} & \textit{non-fl.} &\textbf{all} & \textbf{dist}
& \textit{abs.} & \textit{dep.} & \textit{non-fl.} & \textbf{all} & \textbf{diff}
& \textbf{dist} \\
\midrule

Human
& 8.47 & 4.18 & 10.95 & 23.60 & 0.00
& 16.90 & 9.34 & 16.18 & 34.47 & 0.00
& 0.00 \\
\midrule

llama3.1-8b-i
& \cellcolor{psblue}8.90 & \cellcolor{psblue}3.01 & \cellcolor{psblue}5.08 & \cellcolor{psblue}16.99 & \cellcolor{psblue}28.02
& \cellcolor{psblue}73.21 & \cellcolor{psblue}39.64 & \cellcolor{psblue}56.14 & \cellcolor{psblue}92.35 & \cellcolor{psblue}57.88
& \cellcolor{psblue}42.95 \\
& \cellcolor{rdorange}3.70 & \cellcolor{rdorange}2.25 & \cellcolor{rdorange}0.34 & \cellcolor{rdorange}6.30 & \cellcolor{rdorange}73.31
& \cellcolor{rdorange}52.57 & \cellcolor{rdorange}41.28 & \cellcolor{rdorange}7.93 & \cellcolor{rdorange}72.51 & \cellcolor{rdorange}38.03
& \cellcolor{rdorange}55.67 \\
\midrule
llama3.3-70b-i
& \cellcolor{psblue}7.15 & \cellcolor{psblue}2.79 & \cellcolor{psblue}2.85 & \cellcolor{psblue}12.79 & \cellcolor{psblue}45.79
& \cellcolor{psblue}68.15 & \cellcolor{psblue}40.40 & \cellcolor{psblue}38.60 & \cellcolor{psblue}85.94 & \cellcolor{psblue}51.46
& \cellcolor{psblue}48.63 \\
& \cellcolor{rdorange}6.32 & \cellcolor{rdorange}2.11 & \cellcolor{rdorange}1.04 & \cellcolor{rdorange}9.47 & \cellcolor{rdorange}59.85
& \cellcolor{rdorange}57.16 & \cellcolor{rdorange}27.71 & \cellcolor{rdorange}14.39 & \cellcolor{rdorange}69.91 & \cellcolor{rdorange}35.44
& \cellcolor{rdorange}47.65 \\
\midrule
qwen3-30b-a3b-i
& \cellcolor{psblue}6.68 & \cellcolor{psblue}4.08 & \cellcolor{psblue}11.52 & \cellcolor{psblue}22.28 & \cellcolor{psblue}5.60
& \cellcolor{psblue}49.32 & \cellcolor{psblue}35.36 & \cellcolor{psblue}80.31 & \cellcolor{psblue}92.94 & \cellcolor{psblue}58.47
& \cellcolor{psblue}32.03 \\
& \cellcolor{rdorange}4.91 & \cellcolor{rdorange}2.88 & \cellcolor{rdorange}2.09 & \cellcolor{rdorange}9.88 & \cellcolor{rdorange}58.14
& \cellcolor{rdorange}53.98 & \cellcolor{rdorange}37.15 & \cellcolor{rdorange}25.58 & \cellcolor{rdorange}78.03 & \cellcolor{rdorange}43.56
& \cellcolor{rdorange}50.85 \\
\midrule

qwen2.5-72b-i
& \cellcolor{psblue}5.03 & \cellcolor{psblue}3.95 & \cellcolor{psblue}0.18 & \cellcolor{psblue}9.16 & \cellcolor{psblue}61.20
& \cellcolor{psblue}31.76 & \cellcolor{psblue}28.31 & \cellcolor{psblue}1.49 & \cellcolor{psblue}49.75 & \cellcolor{psblue}15.28
& \cellcolor{psblue}38.24 \\
& \cellcolor{rdorange}4.74 & \cellcolor{rdorange}3.86 & \cellcolor{rdorange}0.03 & \cellcolor{rdorange}8.63 & \cellcolor{rdorange}63.43
& \cellcolor{rdorange}35.51 & \cellcolor{rdorange}34.00 & \cellcolor{rdorange}0.27 & \cellcolor{rdorange}56.11 & \cellcolor{rdorange}21.63
& \cellcolor{rdorange}42.53 \\
\midrule
gpt-4.1-mini
& \cellcolor{psblue}6.95 & \cellcolor{psblue}4.64 & \cellcolor{psblue}0.89 & \cellcolor{psblue}12.49 & \cellcolor{psblue}47.10
& \cellcolor{psblue}57.93 & \cellcolor{psblue}44.98 & \cellcolor{psblue}10.24 & \cellcolor{psblue}78.93 & \cellcolor{psblue}44.46
& \cellcolor{psblue}45.78 \\
& \cellcolor{rdorange}5.02 & \cellcolor{rdorange}4.21 & \cellcolor{rdorange}0.10 & \cellcolor{rdorange}9.33 & \cellcolor{rdorange}60.46
& \cellcolor{rdorange}42.38 & \cellcolor{rdorange}42.04 & \cellcolor{rdorange}1.22 & \cellcolor{rdorange}67.28 & \cellcolor{rdorange}32.80
& \cellcolor{rdorange}46.63 \\
\midrule
gpt-oss-20b
& \cellcolor{psblue}4.63 & \cellcolor{psblue}3.97 & \cellcolor{psblue}1.56 & \cellcolor{psblue}10.16 & \cellcolor{psblue}56.96
& \cellcolor{psblue}45.27 & \cellcolor{psblue}42.30 & \cellcolor{psblue}17.68 & \cellcolor{psblue}72.33 & \cellcolor{psblue}37.86
& \cellcolor{psblue}47.41 \\
& \cellcolor{rdorange}2.68 & \cellcolor{rdorange}2.85 & \cellcolor{rdorange}0.06 & \cellcolor{rdorange}5.58 & \cellcolor{rdorange}76.34
& \cellcolor{rdorange}51.26 & \cellcolor{rdorange}56.76 & \cellcolor{rdorange}1.54 & \cellcolor{rdorange}78.86 & \cellcolor{rdorange}44.38
& \cellcolor{rdorange}60.36 \\
\midrule

gpt-oss-120b
& \cellcolor{psblue}5.17 & \cellcolor{psblue}3.93 & \cellcolor{psblue}4.98 & \cellcolor{psblue}14.08 & \cellcolor{psblue}40.32
& \cellcolor{psblue}52.35 & \cellcolor{psblue}45.81 & \cellcolor{psblue}49.89 & \cellcolor{psblue}84.22 & \cellcolor{psblue}49.75
& \cellcolor{psblue}45.03 \\
& \cellcolor{rdorange}3.36 & \cellcolor{rdorange}3.08 & \cellcolor{rdorange}0.12 & \cellcolor{rdorange}6.56 & \cellcolor{rdorange}72.19
& \cellcolor{rdorange}68.77 & \cellcolor{rdorange}69.25 & \cellcolor{rdorange}4.28 & \cellcolor{rdorange}90.01 & \cellcolor{rdorange}55.54
& \cellcolor{rdorange}63.86 \\

\bottomrule
\end{tabular}

\caption{\textbf{Message-level prevalence and token-normalized rates of depression markers.} \textit{abs.}: absolutist words; \textit{dep.}: depressive words; \textit{non-fl.}: non-fluencies (see lexicon in Table~\ref{tab:depression_marker_lexicon}).
\textit{Marker Rate}: number of marker occurrences per 1{,}000 tokens; \textit{Marker Prevalence}: percentage of messages containing at least one marker. Blue rows indicate Patient-$\Psi$, orange rows indicate Roleplay-doh.}
\label{tab:depression_category_scores}
\end{table*}

\begin{table*}[h]
\centering
\small
\begin{tabular}{l|cc|cc|c}
\toprule
\textbf{Setting} 
& \multicolumn{2}{c|}{\textbf{Words per Message}} 
& \multicolumn{2}{c|}{\textbf{Words per Sentence}} 
& \textbf{Combined} \\
& Count & Sim & Count & Sim & \textbf{similarity}\\
\midrule
Human 
& 18.24 & 100.0 & 9.16 & 100.0 & 100.0 \\
\midrule

llama3.1-8b-i 
& \cellcolor{psblue}141.92 & \cellcolor{psblue}12.85 
& \cellcolor{psblue}22.74 & \cellcolor{psblue}40.28 
& \cellcolor{psblue}26.57 \\
& \cellcolor{rdorange}215.03 & \cellcolor{rdorange}8.48 
& \cellcolor{rdorange}23.45 & \cellcolor{rdorange}39.06 
& \cellcolor{rdorange}23.77 \\

\midrule
llama3.3-70b-i 
& \cellcolor{psblue}143.96 & \cellcolor{psblue}12.67 
& \cellcolor{psblue}27.94 & \cellcolor{psblue}32.78 
& \cellcolor{psblue}22.72 \\
& \cellcolor{rdorange}122.56 & \cellcolor{rdorange}14.88 
& \cellcolor{rdorange}22.67 & \cellcolor{rdorange}40.41 
& \cellcolor{rdorange}27.64 \\

\midrule
qwen3-30b-a3b-i
& \cellcolor{psblue}80.51 & \cellcolor{psblue}22.65 
& \cellcolor{psblue}15.30 & \cellcolor{psblue}59.86 
& \cellcolor{psblue}41.25 \\
& \cellcolor{rdorange}126.94 & \cellcolor{rdorange}14.37 
& \cellcolor{rdorange}17.65 & \cellcolor{rdorange}51.89 
& \cellcolor{rdorange}33.13 \\

\midrule
qwen2.5-72b-i 
& \cellcolor{psblue}63.91 & \cellcolor{psblue}28.54 
& \cellcolor{psblue}17.63 & \cellcolor{psblue}51.96 
& \cellcolor{psblue}40.25 \\
& \cellcolor{rdorange}81.99 & \cellcolor{rdorange}22.24 
& \cellcolor{rdorange}18.93 & \cellcolor{rdorange}48.40 
& \cellcolor{rdorange}35.32 \\

\midrule
gpt-4.1-mini 
& \cellcolor{psblue}97.84 & \cellcolor{psblue}20.10 
& \cellcolor{psblue}20.38 & \cellcolor{psblue}47.57 
& \cellcolor{psblue}33.84 \\
& \cellcolor{rdorange}92.11 & \cellcolor{rdorange}21.36 
& \cellcolor{rdorange}23.91 & \cellcolor{rdorange}40.54 
& \cellcolor{rdorange}30.95 \\

\midrule
gpt-oss-20b 
& \cellcolor{psblue}108.97 & \cellcolor{psblue}16.73 
& \cellcolor{psblue}24.56 & \cellcolor{psblue}37.30 
& \cellcolor{psblue}27.02 \\
& \cellcolor{rdorange}239.39 & \cellcolor{rdorange}7.62 
& \cellcolor{rdorange}26.04 & \cellcolor{rdorange}35.18 
& \cellcolor{rdorange}21.40 \\

\midrule
gpt-oss-120b 
& \cellcolor{psblue}121.13 & \cellcolor{psblue}15.05 
& \cellcolor{psblue}27.11 & \cellcolor{psblue}33.79 
& \cellcolor{psblue}24.42 \\
& \cellcolor{rdorange}318.90 & \cellcolor{rdorange}5.72 
& \cellcolor{rdorange}27.41 & \cellcolor{rdorange}33.43 
& \cellcolor{rdorange}19.57 \\

\bottomrule
\end{tabular}

\caption{Response length comparison between reference human patients and simulators. Blue rows indicate PATIENT-$\Psi$ and orange rows indicate Roleplay-doh. \textbf{Sim} is log-ratio similarity to average words count (per message or per sentence) of human patients. Higher scores indicate closer alignment with human patient verbosity.}
\label{tab:verbosity_scores}
\end{table*}

\begin{table*}[t]
\centering
\small
\setlength{\tabcolsep}{2pt} 
\renewcommand{\arraystretch}{1.0}
\begin{tabular}{cc|l|ccccc|c}
\toprule
\textbf{Rank} Run 1 & \textbf{Rank} Run 2 & \textbf{Setting} &
\textbf{NEP} $\uparrow$ & \textbf{Valence} $\uparrow$ &
\textbf{Div.} $\uparrow$ & \textbf{Length} $\uparrow$ &
\textbf{Markers} $\uparrow$ & \textbf{Overall} $\uparrow$ \\
\midrule
\rowcolor{psblue} 1  & 1  & PS-\texttt{llama3.1-8b-i}
& \textbf{69.92} & \textbf{80.32} & 72.13 & 27.99 & 56.12 & 61.30 \\

\rowcolor{psblue} 2  & 2  & PS-\texttt{qwen3-30b-a3b-i}
& 54.90 & 68.89 & 62.94 & \textbf{43.96} & \textbf{67.95} & 59.73 \\

\rowcolor{psblue} 3  & 3  & PS-\texttt{qwen2.5-72b-i}
& 56.53 & 67.48 & 56.63 & \underline{42.73} & \underline{61.94} & 57.06 \\

\rowcolor{rdorange} 4  & 4  & RD-\texttt{llama3.3-70b-i}
& 56.31 & 66.70 & \underline{72.87} & 29.36 & 52.08 & 55.46 \\

\rowcolor{psblue} 5  & 5  & PS-\texttt{llama3.3-70b-i}
& 58.35 & 71.85 & 70.07 & 24.21 & 51.53 & 55.20 \\

\rowcolor{rdorange} 6  & 6  & RD-\texttt{llama3.1-8b-i}
& 51.97 & 65.76 & \textbf{73.38} & 25.32 & 44.28 & 52.14 \\

\rowcolor{rdorange} 7  & 7  & RD-\texttt{qwen2.5-72b-i}
& 51.78 & 66.50 & 44.32 & 37.65 & 57.30 & 51.51 \\

\rowcolor{psblue} 8  & 8  & PS-\texttt{gpt-4.1-mini}
& \underline{59.71} & \underline{76.59} & 32.45 & 33.81 & 54.22 & 51.35 \\

\rowcolor{psblue} 9  & 9  & PS-\texttt{gpt-oss-20b}
& 53.36 & 68.69 & 42.96 & 28.62 & 52.83 & 49.29 \\

\rowcolor{rdorange} 10 & 10 & RD-\texttt{qwen3-30b-a3b-i}
& 46.57 & 62.29 & 46.75 & 35.78 & 49.15 & 48.11 \\

\rowcolor{psblue} 11 & 11 & PS-\texttt{gpt-oss-120b}
& 55.41 & 74.24 & 21.19 & 25.87 & 54.60 & 46.26 \\

\rowcolor{rdorange} 12 & 12 & RD-\texttt{gpt-4.1-mini}
& 53.53 & 68.78 & 19.26 & 31.01 & 53.26 & 45.17 \\

\rowcolor{rdorange} 13 & 13 & RD-\texttt{gpt-oss-20b}
& 42.85 & 62.44 & 24.36 & 22.81 & 40.00 & 38.49 \\

\rowcolor{rdorange} 14 & 14 & RD-\texttt{gpt-oss-120b}
& 45.68 & 63.69 & 0.00 & 20.80 & 36.36 & 33.31 \\
\bottomrule
\end{tabular}
\caption{Benchmark results for a second full generation run under
settings identical to the main benchmark (Table~\ref{tab:overall_benchmark}).
Rank Run 1 and Rank Run 2 denote the rankings
obtained in the main run and in this repeated run. The two runs
produce an \textbf{identical ordering} across all 14 settings, with overall scores differing by at most 0.50 points and individual dimension scores by at most 1.47 points. Higher scores indicate greater alignment with reference human patient
behavior. PS--: \textsc{Patient}-$\Psi$; RD--: Roleplay-doh; Div.:
lexical diversity. Bold and underline denote the best and second-best
results, respectively.}
\label{tab:rerun}
\label{tab:overall_benchmark_run2}
\end{table*}

\begin{table*}[t]
\centering
\small
\setlength{\tabcolsep}{2pt} 
\renewcommand{\arraystretch}{1.0}

\begin{tabular}{cc|l|ccccc|c}
\toprule
\textbf{Rank }\texttt{gpt} & \textbf{Rank }\texttt{Psy} & \textbf{Setting} &
\textbf{NEP} $\uparrow$ & \textbf{Valence} $\uparrow$ &
\textbf{Div.} $\uparrow$ & \textbf{Length} $\uparrow$ &
\textbf{Markers} $\uparrow$ & \textbf{Overall} $\uparrow$ \\
\midrule
\rowcolor{psblue} 1  & 1  & PS-\texttt{llama3.1-8b-i} & \textbf{71.95} & \textbf{85.98} & 75.82 & 26.49 & 55.85 & 63.22 \\
\rowcolor{psblue} 2  & 2  & PS-\texttt{qwen3-30b-a3b-i} & 58.70 & 69.67 & 68.70 & \textbf{41.53} & \textbf{66.90} & 61.10 \\
\rowcolor{psblue} 5  & 3  & PS-\texttt{llama3.3-70b-i} & \underline{60.20} & 74.01 & 75.46 & 23.75 & 50.89 & 56.86 \\
\rowcolor{rdorange} 4  & 4  & RD-\texttt{llama3.3-70b-i} & 58.66 & 67.83 & \underline{76.79} & 28.96 & 51.96 & 56.84 \\
\rowcolor{psblue} 3  & 5  & PS-\texttt{qwen2.5-72b-i} & 56.20 & 64.88 & 59.50 & \underline{40.96} & \underline{60.28} & 56.36 \\
\rowcolor{psblue} 8  & 6  & PS-\texttt{gpt-4.1-mini} & 60.00 & \underline{74.98} & 43.86 & 33.32 & 54.50 & 53.33 \\
\rowcolor{rdorange} 6  & 7  & RD-\texttt{llama3.1-8b-i} & 53.33 & 64.17 & \textbf{77.27} & 23.34 & 41.45 & 51.91 \\
\rowcolor{rdorange} 7  & 8  & RD-\texttt{qwen2.5-72b-i} & 51.01 & 63.35 & 45.12 & 33.93 & 53.62 & 49.41 \\
\rowcolor{psblue} 9  & 9  & PS-\texttt{gpt-oss-20b} & 55.09 & 66.28 & 42.91 & 27.28 & 52.26 & 48.76 \\
\rowcolor{rdorange} 10 & 10 & RD-\texttt{qwen3-30b-a3b-i} & 46.66 & 60.07 & 45.71 & 31.89 & 48.14 & 46.49 \\
\rowcolor{psblue} 11 & 11 & PS-\texttt{gpt-oss-120b} & 55.14 & 68.17 & 22.52 & 23.83 & 55.17 & 44.97 \\
\rowcolor{rdorange} 12 & 12 & RD-\texttt{gpt-4.1-mini} & 53.40 & 64.92 & 23.12 & 29.15 & 50.43 & 44.20 \\
\rowcolor{rdorange} 13 & 13 & RD-\texttt{gpt-oss-20b} & 42.86 & 60.03 & 14.34 & 22.88 & 35.47 & 35.11 \\
\rowcolor{rdorange} 14 & 14 & RD-\texttt{gpt-oss-120b} & 45.72 & 60.78 & 0.00 & 20.75 & 33.90 & 32.23 \\
\bottomrule
\end{tabular}
\caption{Benchmark results across all dimensions and average overall score for using \texttt{PsyCoPref-Llama3-8B} \cite{zhang2026preference} as a counsellor LLM. Rank \texttt{gpt} and Rank \texttt{Psy} denote rankings obtained using \texttt{gpt-oss-20b} and \texttt{PsyCoPref-Llama3-8B} as counsellor LLMs. Results are highly consistent across two counsellor LLMs: broader ranking patterns are preserved, while closely scored settings (within 2 points) may shift by up to two positions. The top-performing setting for each metric is unchanged. Higher scores indicate greater alignment with reference human patient behavior. \textit{PS--}: PATIENT-$\Psi$; \textit{RD--}: Roleplay-doh; \textit{Div.}: lexical diversity. \textbf{Bold} and \underline{underline} denote the best and second-best results, respectively.}
\label{tab:overall_benchmark2}
\end{table*}

\begin{figure*}[h]
\centering
\small

\begin{itemize}
    \item \textbf{Situation of the Client:} \\
    \textit{The client feels better due to the contemplation of suicide as a solution to their depression, believing it would end their pain and alleviate the perceived burden they impose on their family and friends.}
    \item \textbf{Symptom Severity:}
    \begin{itemize}
        \item Feelings of sadness, tearfulness, emptiness, or hopelessness: \textit{Severe}
        \item Loss of interest or pleasure in most or all normal activities, such as sex, hobbies, or sports: \textit{Moderate}
        \item Feelings of worthlessness or guilt, fixating on past failures or self-blame: \textit{Severe}
        \item Frequent or recurrent thoughts of death, suicidal thoughts, suicide attempts, or suicide: \textit{Severe}
        \item Becoming withdrawn, negative, or detached: \textit{Moderate}
        \item Isolating from family and friends: \textit{Moderate}
    \end{itemize}
    \item \textbf{Cognition Distortion Exhibition:}
    \begin{itemize}
        \item Selective abstraction: \textit{Exhibited}
        \item Catastrophic thinking: \textit{Exhibited}
        \item Minimisation: \textit{Exhibited}
    \end{itemize}

    \item \textbf{Resistance Toward the Support:} \textit{Medium}
    \item \textbf{Depression severity:}\\ \textit{Severe Depression. The client exhibits severe feelings of hopelessness, worthlessness, persistent thoughts of suicide, and a belief that their death would benefit loved ones.}
    \item \textbf{Suicidal ideation severity: }\\ \textit{Severe suicidal ideation. The client has a specific plan to end their life and believes it will resolve their depression and relieve their loved ones.}
    \item \textbf{Homicidal ideation severity:} \\ \textit{No homicidal ideation. There are no indications of thoughts or plans to harm others.}
\item \textbf{Counseling History:} \\
    \textit{The client, Jill, has been feeling depressed and previously had sessions with the supporter. Recently, she has made a plan to commit suicide by taking pills and feels that doing so is the solution to her problems. Despite her determination, she seeks the counselor's help because part of her is unsure and doesn’t want to harm her family.}

\end{itemize}

\caption{An example of patient profile}
\label{fig:psych_profile_jill}
\end{figure*}


\begin{table*}[t]
\centering
\small
\begin{tabular}{llll}
\toprule
\textbf{Family} & \textbf{Alias} & \textbf{Full name} & \textbf{Citation} \\
\midrule
\multirow{3}{*}{OpenAI}
    & gpt-4.1-mini
    & GPT-4.1 mini
    & N/A \\
    & gpt-oss-20b
    & gpt-oss-20b
    & \citep{openai2025gptoss120bgptoss20bmodel} \\
    & gpt-oss-120b
    & gpt-oss-120b
    & \citep{openai2025gptoss120bgptoss20bmodel} \\
\midrule
\multirow{2}{*}{Llama}
    & llama3.1-8b-i
    & Llama-3.1-8B-Instruct
    & \cite{grattafiori2024llama} \\
    & llama3.3-70b-i
    & Llama-3.3-70B-Instruct
    & \cite{grattafiori2024llama} \\
\midrule
\multirow{2}{*}{Qwen}
    & qwen3-30b-a3b-i
    & Qwen3-30B-A3B-Instruct-2507
    & \citep{yang2025qwen3} \\
    & qwen2.5-72b-i
    & Qwen2.5-72B-Instruct
    & \citep{qwen2025qwen25technicalreport} \\
\bottomrule
\end{tabular}
\caption{Models used in our experiments.}
\label{tab:model_alias}
\end{table*}

\begin{table*}[h]
\centering
\begin{tabular}{p{2.5cm} p{8cm} p{2.2cm}}
\hline
\textbf{Category} & \textbf{Terms} & \textbf{Source} \\
\hline
\addlinespace
Absolutist words &
absolutely, all, always, complete, completely, constant, constantly, definitely, entire, ever, every, everyone, everything, full, must, never, nothing, totally, whole &
\cite{doi:10.1177/2167702617747074} \\
\addlinespace
Depressive words &
depression, collapse, stress, suicide, apastia, anxious, sad, tired, death, lonely, insomnia, bad, desperate, give up, low, leave, fear, danger, close, sensitive, lost, shadow, destroy, suspect, crash, dark, helpless, guilt, negative, frustration, nervous, melancholy, rubbish, jump, forget, cut wrist, edge, haze, antidepressant & 
\cite{info:doi/10.2196/17650} \\
\addlinespace
Non-fluencies &  ...,
uh, um, er, ah, eh, oh, hmm, mm, hm, huh, mmm, mhm, you know, y'know, i mean, let's see
 &
\cite{UCAM-CL-TR-915} \\
\hline
\end{tabular}
\caption{Lexicon of Linguistic Markers of Depression used in analysis. The non-fluencies lexicon comprises \textit{filled pauses} and \textit{discourse markers} from \cite{UCAM-CL-TR-915}, expanded with additional terms identified through our observations.}
\label{tab:depression_marker_lexicon}
\end{table*}

\begin{table*}[t]
\centering
\small
\begin{tabular}{llp{2.6cm}}
\toprule
\textbf{Attribute} & \textbf{Category} & \textbf{Count / Summary} \\
\midrule
Age &  & 21--63 \\
 &  & \hspace{1em}\emph{Mean} = 35.3 \\
 &  & \hspace{1em}\emph{SD} = 11.4 \\
 &  & \hspace{1em}Median = 33 \\
\midrule
Gender & Female & 11 \\
 & Male & 9 \\
\midrule
Job role & Psychologist & 9 \\
 & Social worker & 6 \\
 & Counsellor (mental health) & 2 \\
 & Clinical psychologist & 2 \\
 & Occupational therapist & 1 \\
\midrule
Country of residence & United States & 4 \\
 & Sweden & 2 \\
 & Spain & 2 \\
    & United Kingdom & 2 \\
 & Germany & 1 \\
 & Canada & 1 \\
 & India & 1 \\
 & Poland & 1 \\
 & Kenya & 1 \\
 & Italy & 1 \\
 & Czech Republic & 1 \\
 & Australia & 1 \\
 & Portugal & 1 \\
 & Greece & 1 \\
\bottomrule
\end{tabular}
\caption{The anonymized demographic statistics reported here summarize all 20 annotators.}
\label{tab:annotator_demographics}
\end{table*}


\begin{table*}[t]
\small
\setlength{\tabcolsep}{4pt}
\renewcommand{\arraystretch}{1.05}
\begin{tabularx}{\textwidth}{>{\RaggedRight\arraybackslash}X >{\RaggedRight\arraybackslash}X}
\toprule
\multicolumn{2}{l}{\textbf{Patient Profile}} \\
\midrule
\multicolumn{2}{>{\RaggedRight\arraybackslash}p{\dimexpr\textwidth-2\tabcolsep\relax}}{
\textbf{situation of the client:} The client is experiencing emotional distress from negative remarks and feels trapped in a living situation with someone who is emotionally abusive. They are concerned about the potential loss of custody of their children due to their mental health diagnosis. 

\textbf{resistance toward the support:} Medium 

\textbf{symptom severity:} feelings of sadness, tearfulness, emptiness, or hopelessness: Severe; angry outbursts, irritability, or frustration, even over small matters: Moderate; loss of interest or pleasure in most or all normal activities, such as sex, hobbies, or sports: Moderate; tiredness and lack of energy, so even small tasks take extra effort: Moderate; anxiety, agitation, or restlessness: Moderate; feelings of worthlessness or guilt, fixating on past failures or self-blame: Severe; trouble thinking, concentrating, making decisions, and remembering things: Moderate; becoming withdrawn, negative, or detached: Severe; isolating from family and friends: Severe; inability to meet the responsibilities of work and family or ignoring other important roles: Moderate 

\textbf{cognition distortion exhibition:} selective abstraction: Exhibited; overgeneralizing: Exhibited; personalization: Exhibited; minimisation: Exhibited; arbitrary inference: Exhibited 

\textbf{depression severity:} Moderate Depression: The client is experiencing significant distress and symptoms consistent with major depression but is potentially maintainable with appropriate intervention. 

\textbf{suicidal ideation severity:} No suicidal ideation 

\textbf{homicidal ideation severity:} No homicidal ideation 

\textbf{counseling history:} N/A
} \\
\midrule

\textbf{\underline{Conversation A}} (RD-llama3.3-70b-i) & \textbf{Conversation B} \\
\midrule

\textbf{Therapist:} Welcome back. How are you feeling today as we begin? Is there anything in particular you’d like to explore or focus on during our time together?

\textbf{Patient:} I don't know where to start, everything just feels so messed up. My family's been on my case and I'm just trying to keep my head above water. I feel like I'm losing myself in all of this and I don't know how to get back on track. I'm worried about my kids, I don't want them to get taken away from me because of my own issues. I just feel so stuck and I don't know what to do.

&
\textbf{Therapist:} Thank you for beginning today. How are you feeling right now, and what would you like to explore in our conversation? I'm here to listen and support you as you share whatever comes up.

\textbf{Patient:} I’m feeling pretty drained and a bit anxious right now. I’d like to explore how my recent job change is affecting my sleep and overall mood.
\\
\midrule

\textbf{Therapist:} It sounds like you’re carrying a heavy load of worry and feeling pulled in so many directions. What moments or thoughts stand out as the most overwhelming for you right now? When you think about your children and the fear of losing them, what feelings or images come up? If you could picture a small step toward feeling more grounded, what might that look like?

\textbf{Patient:} I'm just so tired of feeling like I'm not good enough. Every argument with my family makes me feel like I'm failing my kids. I see them happy with someone else, and it tears me apart. I wish I could just have one day without feeling like I'm walking on eggshells, waiting for the next problem to arise. Maybe taking a few minutes for myself each day to breathe and focus on what I need would help, but it feels like a luxury I don't have right now.

&
\textbf{Therapist:} It sounds like the transition is draining your energy and adding a layer of worry. I hear you’re noticing changes in both sleep and mood. Could you share what a typical day at the new job feels like for you, and how that might be impacting your rest?

\textbf{Patient:} A typical day at the new job feels like a constant juggling act. I start around 8:30 am, but by the time I actually get into the office or log on remotely, it’s already 9:00. The first thing I do is sift through a backlog of emails and messages, many of which are urgent and require immediate action. That rush sets a frantic tone for the rest of the day.

After the email storm, I usually have back-to-back meetings—often with clients or partners in different time zones—so I’m hopping between 9:30 and 12:30, then again from 1:00 to 4:00. Each meeting feels like a mini-deadline; I’m constantly drafting follow-ups, adjusting spreadsheets, and trying to anticipate the next question. By 4:30 I’m still in a meeting or stuck in a “catch-up” call, so I only get home around 7:30–8:00.

The environment is also a factor: the office is brightly lit, constantly humming with people. Even when I try to decompress at home, the noise from the building’s HVAC system and the occasional phone rings in the hallway keep my mind on alert mode. I find myself replaying the day in my head as soon as I lie down—thinking, “What if I missed that detail?” or “Did I say that correctly?”—which makes it hard to wind down.

[...]
\\
\bottomrule
\end{tabularx}
\caption{Example 1 of pairwise comparison shown to annotators (Part 1). \underline{Underlined conversation} is produced by higher-ranked simulator according to PSI-Bench. [...] shows the message still continues, but is not shown for brevity.}
\label{tab:expert_pairwise_example_pair0479_1_part1}
\end{table*}

\begin{table*}[t]
\ContinuedFloat
\small
\setlength{\tabcolsep}{4pt}
\renewcommand{\arraystretch}{1.05}
\begin{tabularx}{\textwidth}{>{\RaggedRight\arraybackslash}X >{\RaggedRight\arraybackslash}X}
\toprule

\textbf{Therapist:} It sounds like you’re carrying a lot of self-doubt, and the fear of losing your connection with your kids amplifies that. When you imagine a day where you’re not on eggshells, what does that look like for you? What small moments of calm, even if they’re just a few minutes, have you noticed that might still be possible in your current routine?

\textbf{Patient:} When I think about having a day where everything feels okay, I imagine being able to talk to my family without everything blowing up. I see myself playing with the kids, laughing, and just being with them without all this weight. It's like I'm remembering a dream, but it's what I want. In the mornings, before everyone's up, I sometimes sit outside with coffee, and for a bit, everything's still. It's not a lot, but it's something. When I help the kids with homework or read to them, I can forget about all the stress for a little while. Maybe if I find more of those moments, I'll feel more grounded and less like I'm losing it all.

&
\textbf{Therapist:} It sounds like your days are packed with a lot of moving parts, and that constant shift from one urgent task to the next keeps your mind in a state of alertness even after you’re home. How does that ongoing mental “humming” affect how you feel when you finally try to rest? What moments of the day, if any, bring you a sense of relief or calm, and how might you build on those?

\textbf{Patient:} It’s hard to shake that humming. Even when I’m physically home, my brain keeps replaying the day like a loop. I hear the echo of the last meeting’s questions, the tentative “what if” about the email I sent, and the subtle reminder that there’s still a deadline looming. When I finally try to lie down, that humming becomes a backdrop noise that I can’t quiet out—like a low-grade fan that never stops. It makes it difficult to fall asleep; I usually need a full hour of restless tossing before I do finally nod off. And when I do fall asleep, the humming lingers in my dreams—often as intrusive, fragmented images that make the sleep feel shallow and unsatisfying.

I’ve noticed a few pockets of relief, though, mostly in the early part of the day. After the initial email flood, there is a brief window before the meetings start where I can sit with a cup of tea, take a few deep breaths, and jot down a quick to-do list. That small pause feels grounding; I can outline what needs to be done and feel a sense of control that the rest of the day’s chaos can’t give me. In the evenings, when I finish a long session of “catch-up” calls and have a clear sense that I’ve addressed the most urgent items, I sometimes feel a genuine, if fleeting, sense of relief. I also get a brief moment of calm when I step outside for a quick walk or when I take a shower and let the water’s rhythm reset my nerves.
[...]
\\
\midrule

\multicolumn{2}{>{\RaggedRight\arraybackslash}p{\dimexpr\textwidth-2\tabcolsep\relax}}{
\textbf{Annotator 1 -- A}: The answers in Conversation A seems more like a real patients answers because they are not very detailed and seems to seek feedback from the counsellor. Conversation B has very detailed answers and it seems like the patient is solving everything without any help.

\textbf{Annotator 2 -- A}: I chose Synthetic A because it accurately reflects the patient's profile (family struggles, fear of losing children), whereas Synthetic B completely hallucinates an unrelated corporate persona. Although A is still a bit too articulate compared to the real patient's fragmented style, B's massive, self-diagnosing paragraphs are highly unrealistic for someone with severe depression and emotional distress, making A the clear winner.

\textbf{Annotator 3 -- A}: Based on my own professional experience

\textbf{Annotator 4 -- A}: Some clients will talk like B, but that's extremely rare. B was way too descriptive, and used language that most clients don't use. A seemed much more realistic and natural. Most people need to be probed to give further info as well, which is more similar to how A was responding. And this is already implied, but B gave answers that were longer than typical. A just seemed to be more like a real convo. B isn't outrageous, but just less statistically likely

\textbf{Annotator 5 -- A}: A seems much more human, B seems slightly robotic so to speak.
} \\
\bottomrule
\end{tabularx}
\caption{Example 1 of pairwise comparison shown to annotators (continued) and reasoning for their choice. [...] shows the message still continues, but is not shown for brevity}
\label{tab:expert_pairwise_example_pair0479_1_part2}
\end{table*}
\begin{table*}[t]
\small
\setlength{\tabcolsep}{4pt}
\renewcommand{\arraystretch}{1.05}
\begin{tabularx}{\textwidth}{>{\RaggedRight\arraybackslash}X >{\RaggedRight\arraybackslash}X}

\toprule
\multicolumn{2}{l}{\textbf{Patient Profile}} \\
\midrule

\multicolumn{2}{>{\RaggedRight\arraybackslash}p{\dimexpr\textwidth-2\tabcolsep\relax}}{

\textbf{occupation:} computer programmer

\textbf{situation of the client:} The client feels exhausted and shattered all the time for about six months, with work becoming a problem due to physical symptoms such as muscle aches and a lack of energy. Despite numerous medical consultations and tests, no diagnosis has been made, which frustrates the client.

\textbf{resistance toward the support:} Medium

\textbf{symptom severity:} 
feelings of sadness, tearfulness, emptiness, or hopelessness: Mild; 
angry outbursts, irritability, or frustration: Mild; 
loss of interest or pleasure: Mild; 
sleep disturbances: Severe; 
tiredness and lack of energy: Severe; 
anxiety, agitation, or restlessness: Mild; 
slowed thinking or movement: Mild; 
trouble concentrating: Mild; 
physical problems (e.g., headaches): Mild; 
withdrawn behavior: Mild; 
difficulty meeting responsibilities: Mild

\textbf{cognition distortion exhibition:} arbitrary inference: Exhibited

\textbf{depression severity:} Mild Depression

\textbf{suicidal ideation severity:} No suicidal ideation

\textbf{homicidal ideation severity:} No homicidal ideation

\textbf{counseling history:} N/A
} \\

\midrule

\textbf{Conversation A} (PS-gpt-oss-120b) & \textbf{\underline{Conversation B}} (PS-Llama3.1-8B) \\
\midrule

\textbf{Therapist:} It sounds like you’re ready to begin. How are you feeling right now? What would you like to explore or share in our time together today?

\textbf{Patient:} I’m... pretty terrible, to be honest. I woke up way later than I’m supposed to after another night of tossing and turning, and I just felt like a total failure because I couldn’t even get out of bed. The whole day feels heavy---I’m scared I’ll mess up my work, angry at myself for being so exhausted, and kind of lonely because nobody seems to get it. I guess I want to talk about why I keep feeling worthless when I can’t be productive and what I’m supposed to do about it.
&
\textbf{Therapist:} Hello, I hear you’re ready to begin. How are you feeling as we start this session? Is there something particular you’d like to bring up first?

\textbf{Patient:} I'm feeling a bit anxious, to be honest. I've been having a really tough time lately, and I'm not sure if I'm doing the right thing by coming here. I've been feeling so overwhelmed with my fatigue and pain, and I just feel like I'm failing at everything. I'm worried that I'm going to fall behind at work and that my boss is going to be on my case about it. I guess I just want to figure out why I'm feeling so stuck and how I can get back on track.
\\
\midrule

\textbf{Therapist:} It sounds like you’re carrying a heavy load of self-criticism right now. How does that feeling of worthlessness show up for you when you’re not meeting your own expectations? What do you think might be happening in your mind when you notice those thoughts?

\textbf{Patient:} When I finally manage to sit up, the first thing that pops into my head is that I’m a failure, that I can’t do anything right. It feels like a weight pressing down on me, making everything seem pointless and... I start wondering if I’m even worth being around because I’m not productive. The anxiety builds, like a knot in my chest, and I get scared that I’ll let my team down or become a burden. I end up staying in bed longer, scrolling on my phone, and just replaying the same “I’m worthless” loop over and over. It’s exhausting, but I can’t seem to stop the thoughts once they start.

&

\textbf{Therapist:} It sounds like you’re carrying a lot of pressure right now---fatigue, pain, and worries about work all piling up. How does that feeling of being “stuck” show up for you in your daily routine? What do you think might help you feel a bit more in control of those moments when everything feels overwhelming?

\textbf{Patient:} It's just so hard to get started in the morning, even though I know I need to. I'll lie in bed for a while, thinking about all the things I need to do, and how I just can't do them because I'm too tired. Sometimes I'll take a nap, hoping that it'll give me a boost, but it never really does. I feel like I'm just going through the motions, you know? I'm not really living, I'm just existing, and that's a really scary thought.

\\
\midrule
\end{tabularx}

\caption{Example 2 of pairwise comparison shown to annotators (Part 1). \underline{Underlined conversation} is produced by higher-ranked simulator according to PSI-Bench.}
\label{tab:example_part1}
\end{table*}

\begin{table*}[t]
\ContinuedFloat
\small
\setlength{\tabcolsep}{4pt}
\renewcommand{\arraystretch}{1.05}
\begin{tabularx}{\textwidth}{>{\RaggedRight\arraybackslash}X >{\RaggedRight\arraybackslash}X}

\toprule
\textbf{Therapist:} It sounds like that cycle of self-criticism feels almost automatic, almost like a default setting that pulls you into a heavier mood. When those thoughts surface, what sensations or images come with them---aside from the knot in your chest? How would you describe the space between noticing the thought and feeling the pull toward that “failure” narrative?

\textbf{Patient:} I feel this tightness in my throat and my muscles just kind of... ache all over. In my mind I see a picture of myself at the desk, everyone staring, looking disappointed. There’s a tiny gap---maybe a second---where I notice the thought, but then the feeling rushes in like a wave and pulls me straight into that “failure” story. My heart starts pounding and I get a cold sweat. It all happens so fast I barely have a moment to even breathe. 

& 

\textbf{Therapist:} It sounds like the mornings feel like a battle, with fatigue pulling you back even as you know you need to start the day. What small moments, if any, do you notice that bring a brief sense of relief or a spark of motivation? How does the feeling of “existing” rather than “living” show up in your thoughts or actions as the day goes on?

\textbf{Patient:} I think it's the little things, like making a cup of coffee or taking a shower, that give me a tiny bit of energy and make me feel like I'm doing something, even if it's just small. Sometimes, if I'm lucky, I'll get a bit of momentum going and I'll manage to tackle a task or two, but it's always followed by a crash afterwards. I feel like I'm just putting on a mask, pretending to be okay, but deep down I'm still feeling exhausted and defeated. When I'm not working, I feel like I'm wasting my time, like I should be doing something productive, but I'm just too tired to even think about it. I guess it's like I'm trapped in this never-ending cycle of fatigue and guilt, and I don't know how to escape it.
\\

\midrule
\multicolumn{2}{>{\RaggedRight\arraybackslash}p{\dimexpr\textwidth-2\tabcolsep\relax}}{
\textbf{Annotator 1 -- B}: ``B seems most human and the realest of all the conversation in this block as far as I remember, having the best flow of conversation and seemingly something that could be actually heard when talking with a patient.``

\textbf{Annotator 2 -- A}: ``I chose Synthetic A because it actively incorporates the specific physical and sleep symptoms from the patient profile (e.g., tossing and turning, waking up late, muscle aches all over). It also uses more natural speech patterns, like hesitation and fragmented thoughts. Synthetic B is not terrible and correctly identifies the fatigue and work anxiety, but it feels a bit more generic. The patient in B talks in broad clichés (I am not really living, I am just existing, putting on a mask), whereas the patient in A grounds their anxiety in concrete physical sensations (tightness in the throat, muscle aches, pounding heart), which better matches the profile of someone suffering from severe unexplained physical exhaustion.``

\textbf{Annotator 3 -- B}: ``In B it´s not all negative, but a mix of hope and anxiety``}\\
\bottomrule
\end{tabularx}
\caption{Example 2 of pairwise comparison (continued) and reasoning for annotators’ non-tie choices.}
\label{tab:example_part2}
\end{table*}

\end{document}